\documentclass[a4paper,conference]{IEEEtran}

\usepackage{cite}
\usepackage{graphicx}
\usepackage{amsmath}
\usepackage{booktabs} 
\usepackage{multirow}
\usepackage{amssymb}
\usepackage{url}
\usepackage{hyperref}

\begin{document}
%
\title{Explainable Online Validation of Machine Learning Models for Practical Applications}



\author{
    \IEEEauthorblockN{Wolfgang Fuhl\IEEEauthorrefmark{1}, Yao Rong\IEEEauthorrefmark{1}, Thomas Motz\IEEEauthorrefmark{2}, Michael Scheidt\IEEEauthorrefmark{3},\\ Andreas Hartel\IEEEauthorrefmark{2}, Andreas Koch\IEEEauthorrefmark{2}, Enkelejda Kasneci\IEEEauthorrefmark{1}}
    \IEEEauthorblockA{\IEEEauthorrefmark{1}Human-Computer Interaction, University of T\"ubingen, Germany
    \\\{wolfgang.fuhl, yao.rong, enkelejda.kasneci\}@uni-tuebingen.de}
    \IEEEauthorblockA{\IEEEauthorrefmark{2}Liebherr-Elektronik GmbH, Germany
    \\\{thomas.motz, andreas.hartel, andreas.koch \}@liebherr.com}
    \IEEEauthorblockA{\IEEEauthorrefmark{3}Scheidt MessTechnik, Switzerland\\\{michael.scheidt.external\}@liebherr.com}
}

\maketitle

\begin{abstract}
We present a reformulation of the regression and classification, which aims to validate the result of a machine learning algorithm. Our reformulation simplifies the original problem and validates the result of the machine learning algorithm using the training data. Since the validation of machine learning algorithms must always be explainable, we perform our experiments with the kNN algorithm as well as with an algorithm based on conditional probabilities, which is proposed in this work. For the evaluation of our approach, three publicly available data sets were used and three classification and two regression problems were evaluated. The presented algorithm based on conditional probabilities is also online capable and requires only a fraction of memory compared to the kNN algorithm.
\end{abstract}


\IEEEpeerreviewmaketitle

\section{Introduction}
Neural networks, decision trees or support vector machines are used in many industrial areas today. In sensor technology and data analysis these machine learning methods are used to convert a multitude of complex measured values and data into useful information~\cite{naghibi2016gis,alsheikh2014machine,hu2010qelar,mohr2017personal,ICMV2019FuhlW,AAAIFuhlW,032017,0320170,ACTNEURO2017,Bahmani2016}. This includes position and orientation~\cite{wu2010study,liao2010anomaly,sng2002strategy,kuleshov2017mobile}, monitoring of biometric data~\cite{mohr2017personal,holzinger2016interactive,farrar2012structural,char2018implementing,beam2018big,UMUAI2020FUHL}, control of robots~\cite{kuleshov2017mobile,sng2002strategy,dillmann2000learning,klingspor1997human}, malware analysis~\cite{shamili2010malware,chowdhury2017protecting,hu2013mutantx}, credit rating~\cite{tsai2010credit,huang2004credit,lee2007application,luo2017deep}, visualization~\cite{ROIGA2018,ASAOIB2015,DWTE022017,AGAS2018}, eye movement evaluation~\cite{FCDGR2020FUHL,fuhl2018simarxiv,ICMIW2019FuhlW1,ICMIW2019FuhlW2,EPIC2018FuhlW,C2019,FFAO2019}, and much more. The ever-increasing number of applications for machine learning methods is due to the fact that they are very economical to use compared to the classical algorithms which usually demand for a more detailed knowledge about the modeled system and are less generic.
Another advantage of machine learning methods is the steadily improving performance compared to classical algorithms. As a result, new areas where automation through machine learning methods is gaining ground are constantly growing.

A disadvantage of machine learning methods is that they cannot be easily explained and validated~\cite{rodvold1999software,cheng2018neural}. In many areas, where living beings are involved and safety standards must be guaranteed, the application of complex procedures is not possible because their reliable function cannot be proven. Of course, there are exceptions such as the k Nearest Neighbour method (kNN) or probabilistic models such as the Gaussian Mixture Models (GMM). Unfortunately, these methods lack flexibility to work well for all problems and for example in case of the kNN, huge training datasets result in a computationally expensive classifier.

The range of application of an automatic online validation for machine learning algorithms~\cite{ICMV2019FuhlW} is manifold since classical algorithms are already replaced by machine learning algorithms~\cite{WTCDAHKSE122016,WTCDOWE052017,WDTTWE062018,VECETRA2020,CORR2017FuhlW1,ETRA2018FuhlW,ICCVW2019FuhlW,CAIP2019FuhlW,ICCVW2018FuhlW,WTDTWE092016,WTDTE022017,WTE032017,NNETRA2020}. An important point here is of course the validation of the output of the machine learning model. The most common areas of application are online measurement systems~\cite{webster2018measurement,czichos2018measurement} where data is collected via sensors from the areas of environmental~\cite{cross2017use,bibri2018iot}, biological~\cite{bell2016associations,shcherbina2017accuracy} or technical systems~\cite{muller2017organic,serov2017cognitive}. Thus, these measuring systems are used for monitoring, control and long-term data acquisition, although in most cases they are not safety-critical applications. Nevertheless, a validation of the result is desirable, because this way a new run can be carried out or the user can be informed that the result is probably wrong and can thus exclude it for a data analysis. Also, the online validation can indicate defective sensors, because the output is certainly no longer valid for defective sensors. In the case of safety-critical applications it is even mandatory to validate the model and in case of very high safety requirements like a nuclear plant it is even mandatory to prove the functionality of the model.

In this work, we address the validation and explainability of the results of complex machine learning algorithms. For this purpose, we present an additional reformulation of the general problem definition for machine learning methods, which is described in detail in Section \ref{sec:method}. 

Our contributions to the state of the art are as follows:
\begin{description}
	\item[1] Reformulation of the general problem definition for machine learning methods.
	\item[2] Provable and explainable validation for machine learning algorithms in general.
	\item[3] Realization of our proposed validation approach using conditional probabilities and a kNN model with feature-based qualitative metrics.
	\item[4] Empirical evaluation of our approach to classification and regression problems on public data sets.
\end{description}

\section{Related Work}
The first approach for Neural Network validation was published  in \cite{bishop1993neural,bishop1994novelty}. As with our approach, the data is considered as a distribution and a manually defined threshold is used to decide whether the data is valid or not. The biggest difference to our approach is that we do not have to set thresholds, we formulate an additional quality metric per data input value, and we formulate the classes and quantified regression steps as conditional probability. In \cite{krogh1995neural} the authors used the difference in the responses of different neural network architectures trained on different subsets of the training data. If the difference between the model responses was too large, an invalid response can be assumed. A disadvantage of this approach in practical application is that it does not trace back to a cause of the error regarding the input data. A list of restrictions for neural networks and machine learning algorithms in general is given in \cite{kurd2003establishing,kurd2007developing}. One of the main problems that neural networks face is that they often lack a kind of white-box view of behaviour~\cite{kurd2003establishing,kurd2007developing}. A layered validation and verification of online adaptive neural networks was presented in \cite{schumann2003verification}. Here, the input and output of the neural network are monitored and confidence intervals are calculated using input data. If the input data differs too much from the previous seen input data, the output is considered uncertain. Since this work is in progress, as stated by the authors, the exact calculation was not explained in detail. 

The latest approaches to the validation of neural networks attempt to validate every single neuron in a network. This is done by reformulating it into a linear programming formalism which is then checked using box constraints~\cite{bunel2020branch,bunel2018unified}. Theoretically, it is the best approach but since it cannot be used online, only given input and output conditions are tested, and not all neurons can be validated in general, a lot of research is needed before these approaches can be applied in practice. In addition, \cite{bunel2020branch,bunel2018unified} are limited to neural networks.

\section{Method}
\label{sec:method}
In this work, we deal with the validation and explainability of the results of complex machine learning algorithms. For this purpose, we present an additional reformulation of the general problem definition for machine learning methods. In the case of classification, this problem is given by $M(F^n) \rightarrow \mathbb{N}$, where $M$ is the function which is calculated by the machine learning method, $F^n$ is the $n$-dimensional data or sensor values and $\mathbb{N}$ is the class output. To validate the output class $k \in \mathbb{N}$, we suggest a \textit{"backprojection"}, which is a binary classification. Thus the \textit{"backprojection"} can be represented as $V(F^n ,M(F^n)) \rightarrow \{-1,1\}$. This classification indicates whether the output class $M(F^n)$ represents a valid solution for the given input data $F^n$. The advantage of this formulation is that the complex multiclass problem can be mapped to a simpler binary classification problem. Therefore, simple machine learning methods for $V$ can be used without having to expect large losses in the accuracy of the prediction method $M$.

In the case of regression, the underlying problem is $M(F^n) \rightarrow \mathbb{R}^m$. To make our procedure applicable to such a problem, we use quantization on the real output $\mathbb{R}^m$ to form a multiclass classification problem. The formula is $Q(\mathbb{R}^m) \rightarrow \mathbb{N}^m$ where $Q$ is the quantization and each $m$ can have a different division. With this quantization, the \textit{"backprojection"} can be defined as $V(F^n ,Q(M(F^n)) \rightarrow \{-1,1\}^m$. This means that each output value $m$ of the machine learning model is considered either as valid $1$ or invalid $-1$.

Since in real or industrial applications the pure class assignment is usually not sufficient, a quality signal can be determined by the distance to the individual classes ($\{-1,1\}$) and this can also be mapped to the value range $[0,1]$. In the case of a kNN for the backprojection, the distance of the k selected neighbors to the input data would be exactly this distance. Mapped to the value range $[0,1]$, this would allow the statement how similar the input data is to already known data. Thus, the result of the machine learning process can be easily explained using the training data. Since a kNN requires a lot of memory for the individual training samples and is expensive to calculate with increasing amounts of training data, we describe the backprojection by means of a conditional probability in the following. In addition, the individual input data streams are mapped to distributions and thus allow to evaluate the quality of individual data streams in general.

\begin{figure*}
  \centering
  \includegraphics[width=0.8\linewidth]{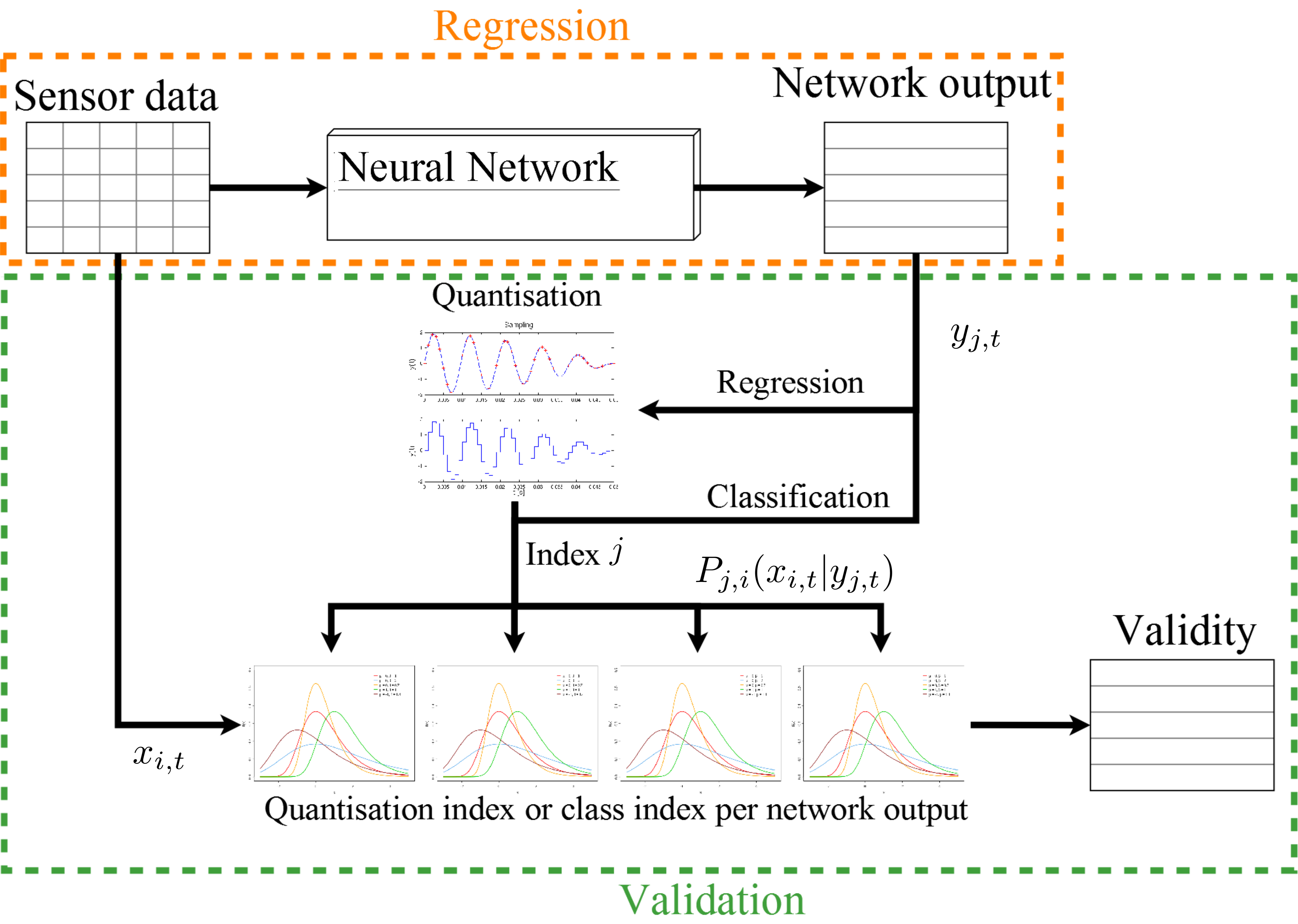}
  \caption{The framework of our proposed online validation method. The orange frame contains the algorithm whose output we want to validate. The green box contains the validation procedure. It consists of an online generation of a conditional probability distribution (Equation~\ref{eq:general}) which is compared with a distribution (Equation~\ref{eq:fit}) generated from a fit against the training dataset during the training of the algorithm. The accordance of both distributions is an indicator for the validity of the algorithm's output and quantitatively estimated by Equation~\ref{eq:validity}.}
  \label{fig:workflow}
\end{figure*}

Figure~\ref{fig:workflow} shows the process of our novel approach for the online validation of the results from machine learning approaches. In Figure~\ref{fig:workflow}, we selected a neural network as an instance. The upper part (orange) is the simple execution. Here the network is applied to sensor data and delivers a result. 

The lower part (green) is our approach to validate the result. In case of a regression, the results of the neural network are quantified to form conditional probability distributions.  This is already given for the classification. Each quantification level or class forms an index to probability distributions calculated on the sensor data. Since these form a high-dimensional vector in most cases, either a dimension reduction must be applied or each individual sensor value must be represented as a probability distribution. Using the quantified result of the network, it is now possible to calculate from the distributions how certain one can be that the input data matches the data already seen (training data) with respect to the result. This gives you the information whether you have already evaluated the net on similar data and the result is the same or not. It makes it possible to validate online whether the sensor values match the data that was used for the approval of the algorithm. Another advantage of the distributions on the sensor data is the quality of the distribution. In high-dimensional input vectors, there may be single values that do not contribute to certain results. Therefore the distribution has no significance or a high degree of uncertainty for the result. This can be measured by the difference of the integrals of the raw data and the distribution. This measure of uncertainty in combination with the assignment to quantified results allows to weight the contribution of individual distributions to the validity differently.

In the following, the whole procedure is described mathematically. During the operational mode, we aim to compute a conditional probability
\begin{equation}
	P^{NET}_{j,i}(x_{i,t}|y_{j,t})
	\label{eq:general}
\end{equation}
which is the probability of $x$ given the output $y$ at time $t$ of the machine learning approach. Since the input raw values $x$ and the output $y$ are vectors, we consider each index $i$ of the input vector $x$ and each position $j$ of the output vector $y$ separately. $y$ contains the discretized sensor values where the index $j$ labels the intervals. The validity of the output is judged by comparison of the distribution in Equation~\ref{eq:general} with a second distribution which was fitted beforehand against the training dataset:

\begin{equation}
P^{GT}_{j,i}(x_{i,t}|y_{j,t})
\label{eq:fit}
\end{equation}

Equation~\ref{eq:fit}, represents the calculated distribution for $y_{j}$, which was fitted using the input data $x_{i}$ and the ground truth labels together with the network output. Therefore, this conditional probability distribution can be understood as a reliability function of the network. This can be done either as histogram or using a fitting algorithm to an formally defined function like a gauss
distribution. The formally defined function itself is freely selectable, but care must be taken that the difference between the output of the formally defined distribution and the raw distribution is as small as possible. We continue with the calculation of the normalized intersection of Equations~\ref{eq:general} and
\ref{eq:fit}:
\begin{equation}
Q_{j,i}=\frac{ P^{NET}_{j,i}(x_{i,t}|y_{j,t}) \cap P^{GT}_{j,i}(x_{i,t}|y_{j,t})}{ (P^{NET}_{j,i}(x_{i,t}|y_{j,t}) \cup P^{GT}_{j,i}(x_{i,t}|y_{j,t})) + \epsilon}
\label{eq:quality}
\end{equation}
 
Equation~\ref{eq:quality} determines the quality of the distribution and corresponds to the Jaccard Index for area comparisons. This means that the more similar the two distributions are, when superimposed, the higher is the quality. For the comparison of distributions $P,Q$ the Kullback-Leibler divergence $D(P,Q)$ is normally used, and is characterized by an asymmetry $D(P,Q) \neq D(Q,P)$. As we have no preferred ordering and store the distributions as histograms, it makes sense to use the symmetric Jaccard Index for our implementation. The normalization leads to a maximum value of 1 and a minimum value equal to 0. $\epsilon$ is needed to prevent a division by 0. An advantage of this formulation is that it is very sensitive to gaps in the distribution and results in a poorly rated quality if the data base is small. With these three equations we can formulate the overall evaluation of new input data $x$ with an output value $y$ at time $t$.
\begin{equation}
V(j,t)= \frac{ \sum_{i=1}^N Q_{j,i} * P^{GT}_{j,i}(x_{i,t}|y_{j,t}) }{ \sum_{i=1}^N \mathrm{max}(P^{GT}_{j,i}(x_{i,t}|y_{j,t}))) }
\label{eq:validity}
\end{equation}

Equation~\ref{eq:validity} describes the computation of the validity of each $j$th response of the machine learning algorithm separately, which is illustrated as neural network in Figure~\ref{fig:workflow}. The computation consist of the quality of the features ($Q_{j,i}$) and the reliability of the network output per feature ($P^{GT}_{j,i}(x_{i,t}|y_{j,t})$). For normalization and to ensure numerical stability, the sum between the product of the quality of the features and the reliability of the network output per feature is divided by the maximum values of the reliability of the network output per feature ($P^{GT}_{j,i}(x_{i,t}|y_{j,t})$) since those values are the maximal outcomes of the product ($Q \in [0,1]$).

For a large validity $V(j,t)$, the scalar product $\sum_{i=1}^N Q_{j,i} * P^{GT}_{j,i}(x_{i,t}|y_{j,t}))$ has to be large, which is the case for a precomputed distribution $P^{GT}_{j,i}(x_{i,t}|y_{j,t})$ that coincides well with the one based on the model data alone $P^{NET}_{j,i}(x_{i,t}|y_{j,t})$. Then, $Q_{j,i} $ and $P^{GT}_{j,i}(x_{i,t}|y_{j,t})$ are similar, which maximizes the numerator in Equation~\ref{eq:validity}. The normalization is done using the sum of maxima of all the included distributions which leads to a preference of homogeneous distributions with small maxima. The validity $V(j,t)$ has a maximum value of 1 and a minimum value of 0. This quality signal can be used to measure whether the input data follows the data used to validate the algorithm or not. Thus, it is possible to make online statements whether the response of the neural network is reliable and has been tested and is therefore explainable and comprehensible. Alternative formulations for Equation~\ref{eq:validity} can use the median of all computed values or predefine a minimum of quality, by using a threshold.

The formulation of Equation~\ref{eq:validity} is limited by the need for sufficiently many data points which correspond to the assigned output and provide a high quality according to Equation~\ref{eq:quality}. If those data is not given, no statement can be made with regard to the input data, since these cannot be represented as a distribution either. 

To circumvent this limitation we follow the principle of divide and conquer. We determine a large number of distributions that fit as well as possible to a local value of the output. This way we simplify the complexity of the whole distribution to many smaller distributions. In addition, we also divide the nonlinear output into ranges (quantization) or use the predefined classes. Since the individual outputs of the machine learning algorithm are also considered separately, the validity can be evaluated separately for each output, but also in total for the whole network (Equation~\ref{eq:all}).

\begin{equation}
V_{all} = \frac{\sum_{j=1}^M V(y_{j})}{M}
\label{eq:all}
\end{equation}

Equation~\ref{eq:all} describes the evaluation of the whole network over all outputs $j$ where the total number of outputs is $M$. This is a simple average where each output can be weighted to obtain a weighted average validity ($V_{all,w} = \frac{\sum_{j=1}^M w_{j} * V(y_{j})}{\sum_{j=1}^M w_{j}}$ with $w_j$ as weight per output).

\section{Evaluation}
In this part of the work, different non-linear machine learning approaches (neural networks, gradient boosted decision trees, and bagged decision trees) are applied to three public data sets. For each machine learning approach we evaluated different model sizes, but only a small part is shown here. The remaining evaluations are in the supplementary material. For the evaluation, we used three classification challenges and two regression challenges. For the validation, we show the results of the both approaches discussed in the last section: The probabilistic approach, and the kNN approach based on the reformulation and histograms as distributions. First, we describe the public data sets used.

\textbf{Beach Water Quality~\footnote[1]{https://data.cityofchicago.org/Parks-Recreation/Beach-Water-Quality-Automated-Sensors/qmqz-2xku}:} This data set contains the sensor data from the water quality of the Chicago Park District along Chicago's Lake Michigan lakefront. It is recorded on six different beaches which are our target classes for the classification experiment. For regression, we used the wave height sensor response. In total it has 39.469 entries but we omitted all records where at least one value was missing. This was done to make all machine learning based approaches like neural networks applicable to the data. Therefore, we used 10.034 records for our evaluation.

\textbf{Classify Gestures by Reading Muscle Activity~\footnote[2]{https://www.kaggle.com/kyr7plus/emg-4}:} This data set was recorded using a MYO armband which records the signals from eight EMG (Electromyography) sensors. Eight of such recordings are connected (64 features per class) and linked to a muscle activity (Gesture classes were : rock - 0, scissors - 1, paper - 2, ok - 3). These gestures are the targets for our classification experiment. We did not use this data set for regression since the sensor values would not make much sense as targets. In total the data set has 11.678 records which we all used in our experiment.

\textbf{Red Wine Quality~\footnote[3]{https://www.kaggle.com/uciml/red-wine-quality-cortez-et-al-2009}:} This data set was published in \cite{cortez2009modeling}. It is related to red variants of the Portuguese "Vinho Verde" wine and contains eleven features like fixed acidity, volatile acidity, citric acid, residual sugar, chlorides, free sulfur dioxide, etc., as well as a numerical quality measure. For classification, we used the numeric quality measure as target. For our regression experiment, we selected citric acid as target value. It is a highly unbalanced data set with 1.599 records. We used all of the records for our evaluation.

\subsection{Evaluated Machine Learning Approaches}
\begin{figure}[htb]
	\centering
	\includegraphics[width=0.8\linewidth]{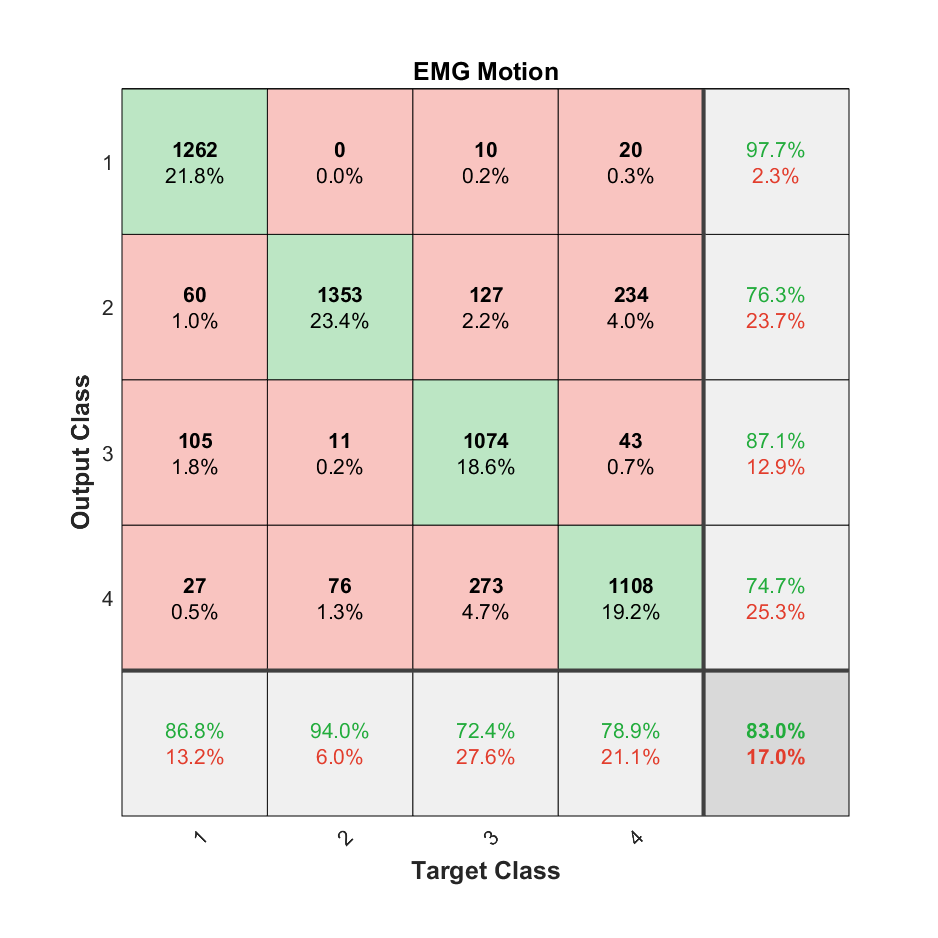}
	
	\includegraphics[width=0.3\linewidth]{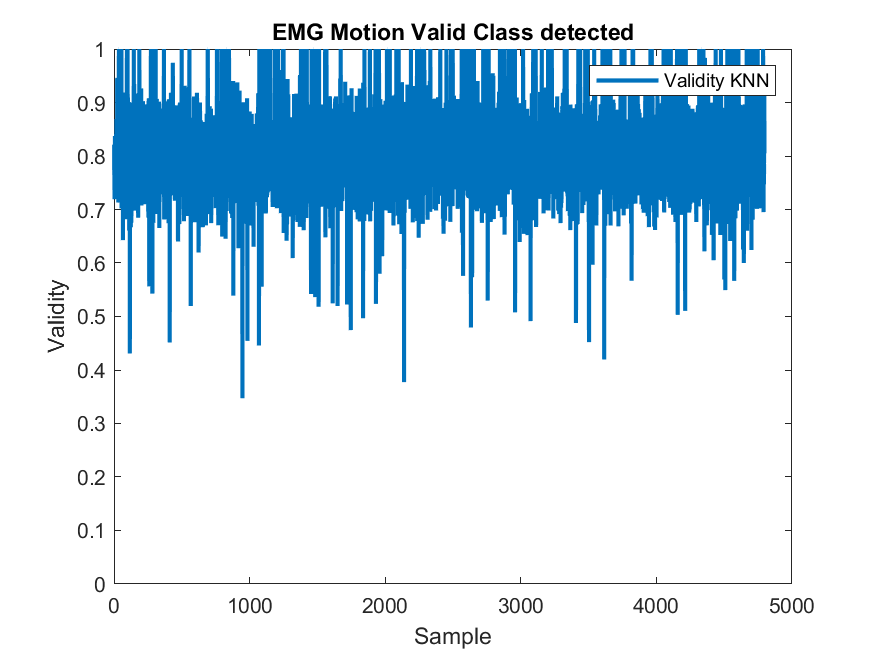}
	\includegraphics[width=0.3\linewidth]{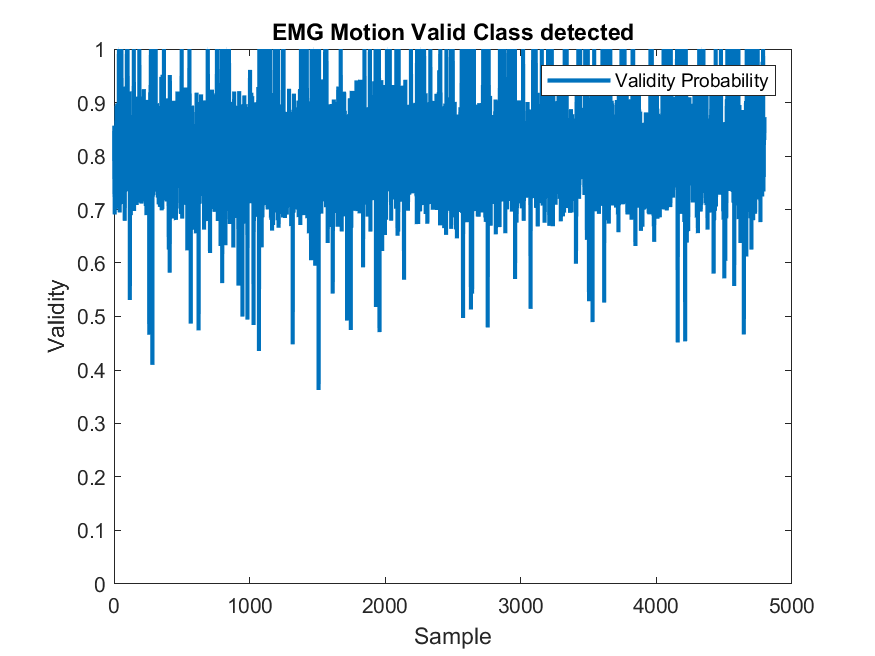}
	\includegraphics[width=0.3\linewidth]{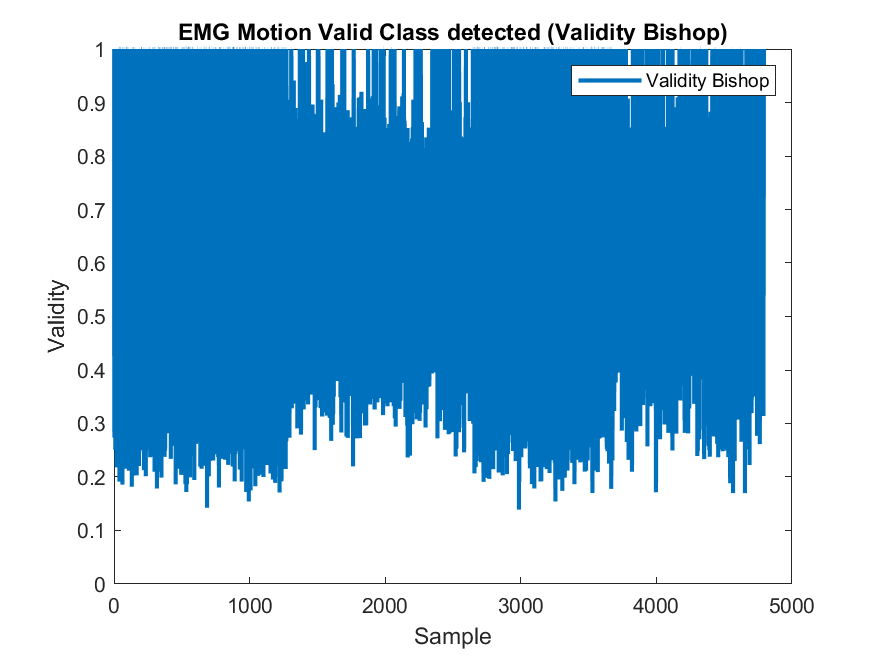}
	
	\includegraphics[width=0.3\linewidth]{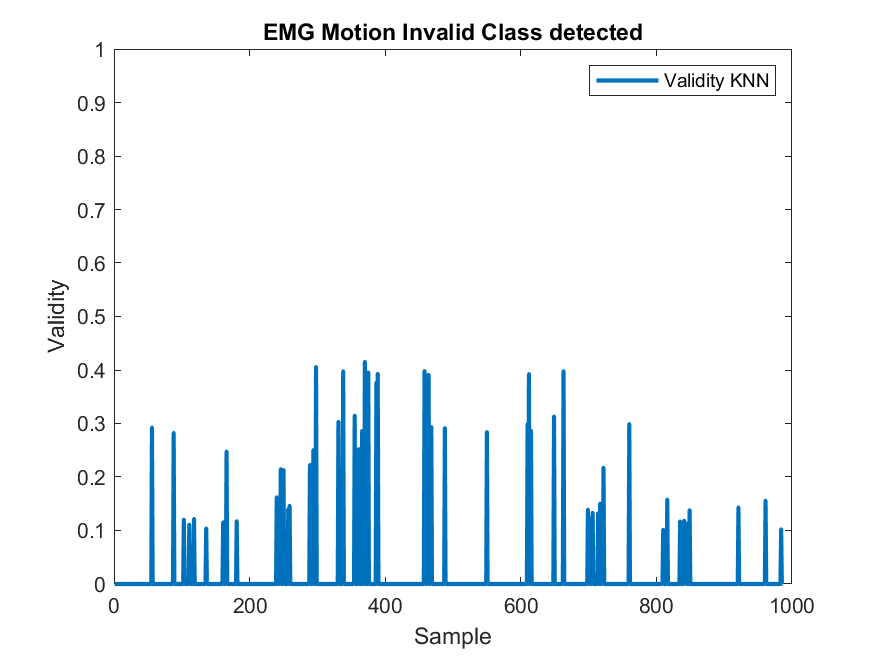}
	\includegraphics[width=0.3\linewidth]{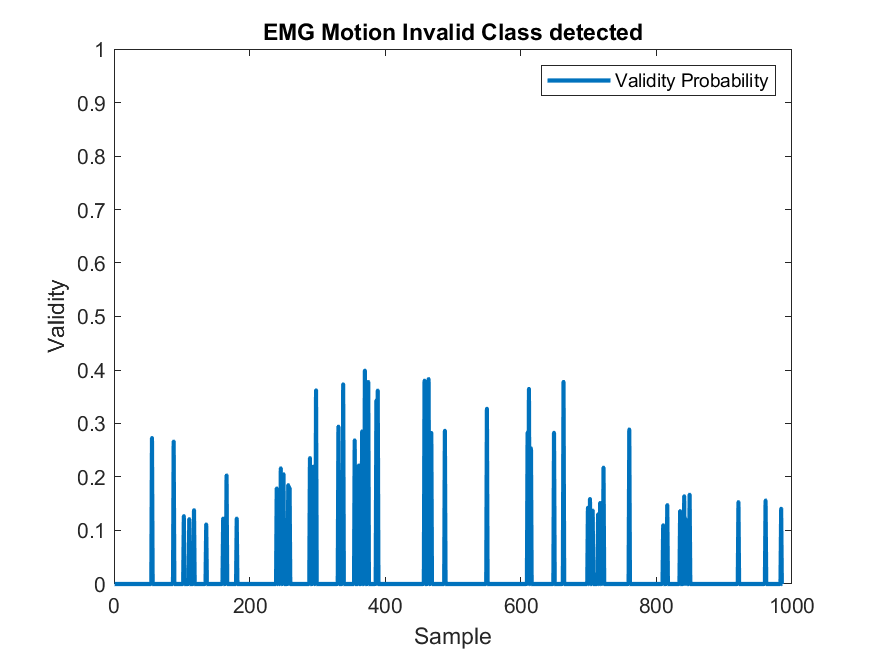}
	\includegraphics[width=0.3\linewidth]{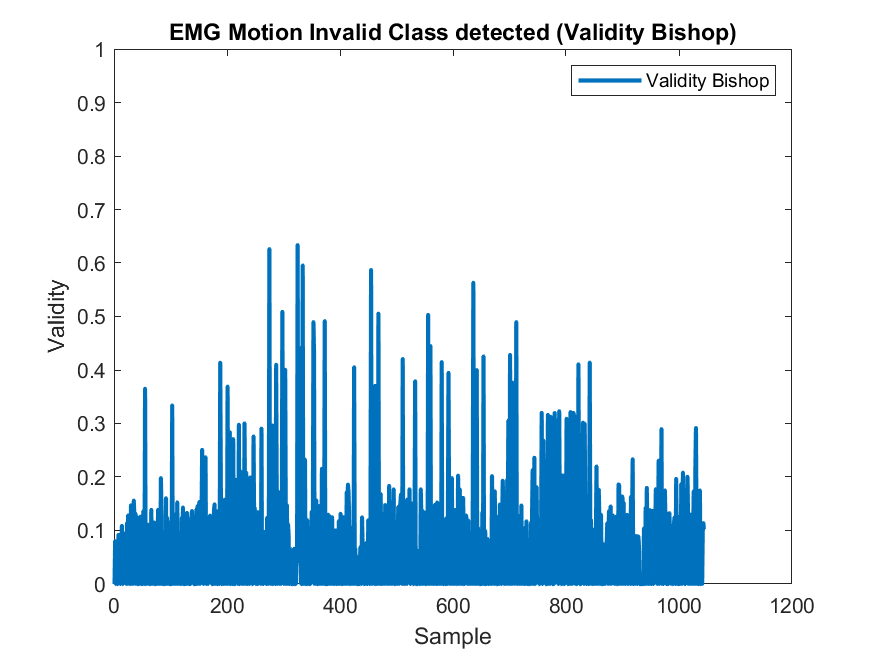}
	\caption{The first (top) visualization is a confusion matrix for the Gestures classifier accuracy. The bottom right corner of the confusion matrix shows the overall accuracy of the classifier. The three central plots  show the validity signal per sample for correctly classified examples (left kNN, center Probability, and right re-implementation of \cite{bishop1993neural}). For the incorrect classified examples the validity signal per sample is shown in the three plots at the bottom in the same order.}
	\label{fig:class1}
\end{figure}
To test our approach extensively, we evaluated three machine learning algorithms in four different configurations each. The selected machine learning algorithms are neural networks, bagged ensemble of decision trees, and gradient boosted ensemble of decision trees. We selected those because all of these approaches are known to perform very well on nearly all data sets. For the neural network, we used 20, 30, 40, and 50 neurons as one hidden layer before the final output neuron. For the regression we used Levenberg Marquart backpropagation~\cite{hagan1994training} for training. For the classification we used
 Scaled Conjugated gradients~\cite{moller1990scaled}. The bagged and boosted ensembles were used with 5, 10, 15, and 20 decision stumps. For the classification with bagging, we used the standard randomized approach form random forests~\cite{breiman2001random} and quantile regression forests~\cite{meinshausen2006quantile} for the regression. The classification with boosting was done using totally corrective boosting~\cite{warmuth2006totally}, and for the regression we used least-squares boosting~\cite{hastie2009elements}.

The validity signal was estimated using a kNN~\cite{hart1968condensed}. As an alternative using our proposed approach without explicit distribution fitting for the features, we used the estimated distribution from the data as histogram. This means that our quality of the distribution is always one (Equation~\ref{eq:quality}). For the kNN we set K as the number of features in the data set and used euclidean as distance metric. In addition, we compared the distance to all known entries instead of an approximation with a tree structure. In addition, we compare our results with the previous work~\cite{bishop1993neural} to show the advantage of our algorithm.

The train and test split was done using a 50\% to 50\% split. This means we used 50\% for training of the classifier and for the validation of them. This data was also used to train the validation algorithms. The other 50\% are for evaluation only. 

Since showing the complete evaluation of all the models for three classifications and two regression problems would exceed the content of this work, only the results of the neural network with 50 neurons in the hidden layer were included. All other results are found in the supplementary material.

\subsection{Results}
\begin{figure}[htb]
	\centering
	\includegraphics[width=0.8\linewidth]{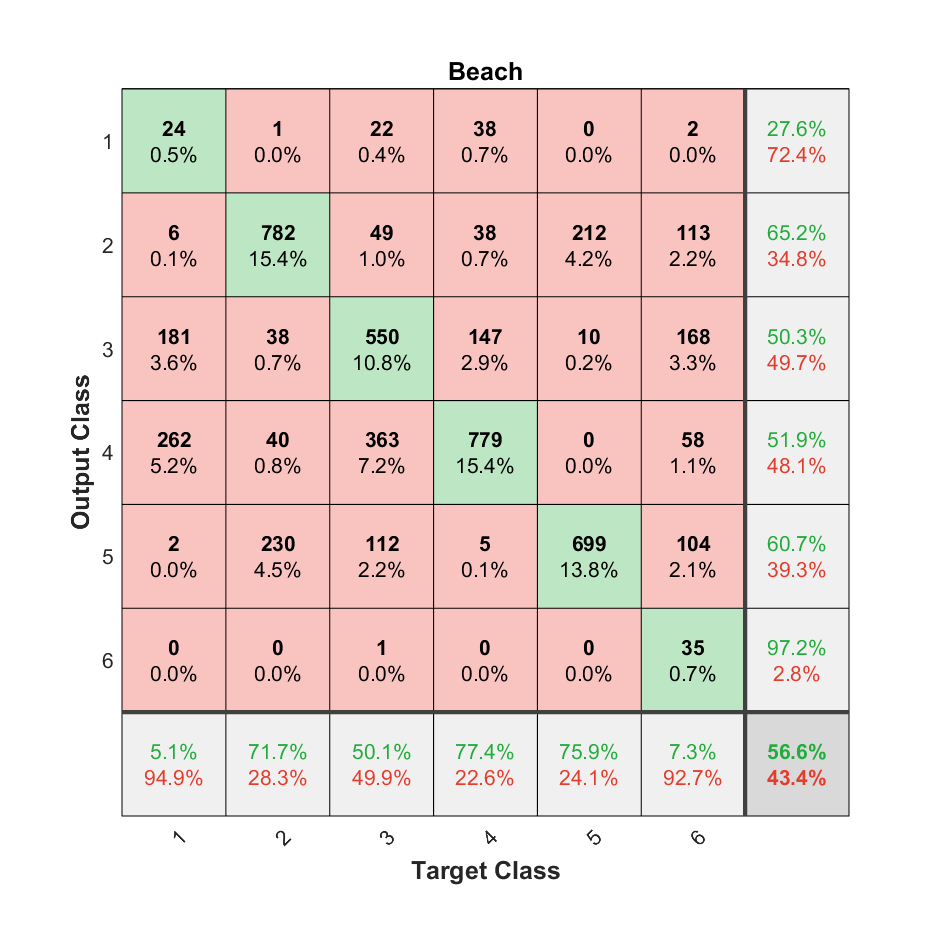}
	
	\includegraphics[width=0.3\linewidth]{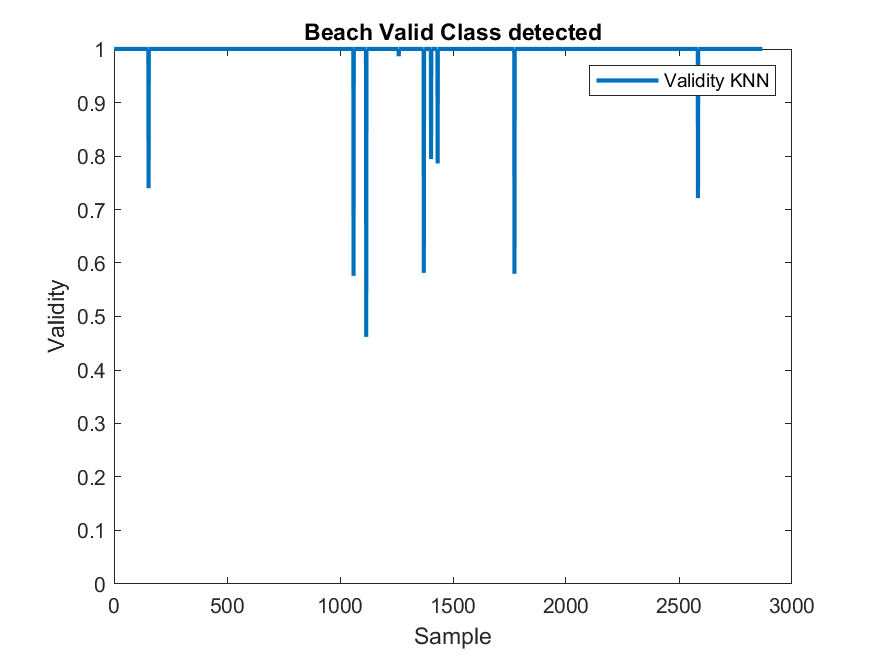}
	\includegraphics[width=0.3\linewidth]{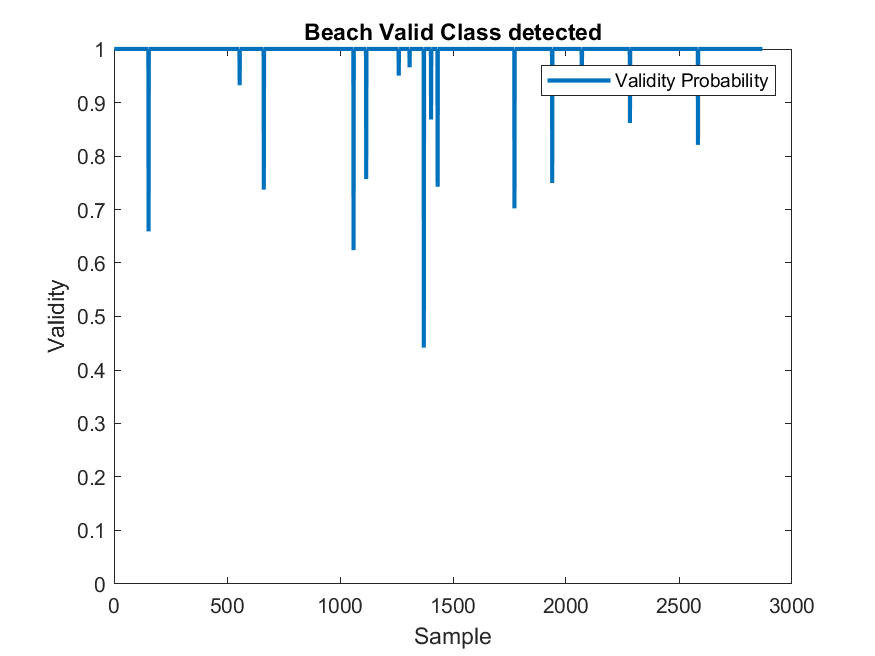}
	\includegraphics[width=0.3\linewidth]{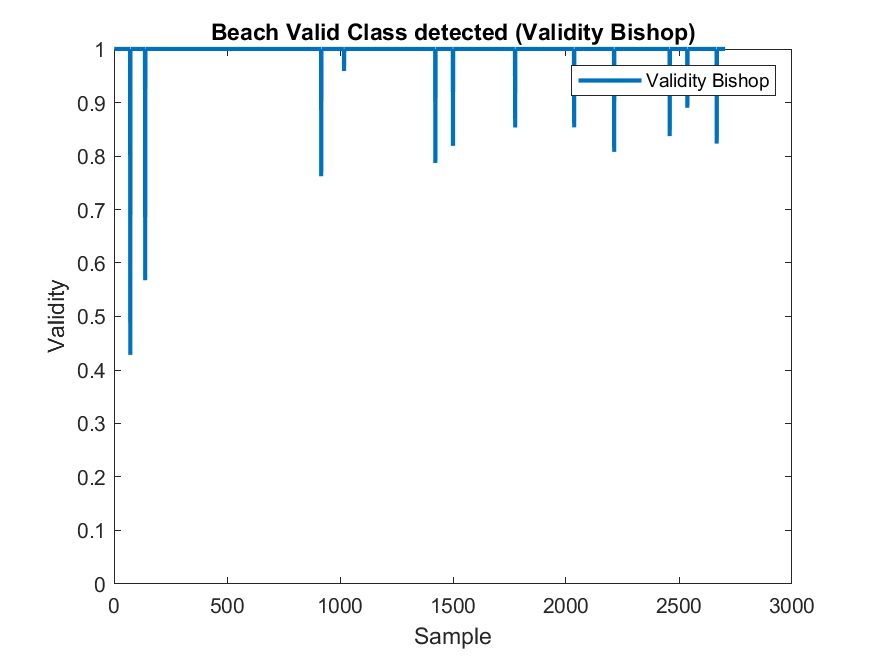}
	
	\includegraphics[width=0.3\linewidth]{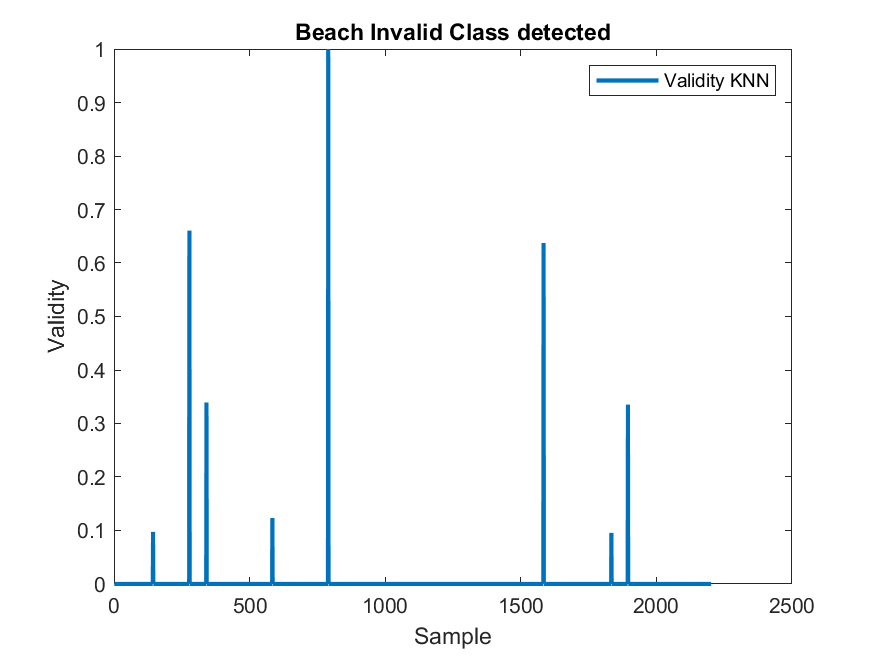}
	\includegraphics[width=0.3\linewidth]{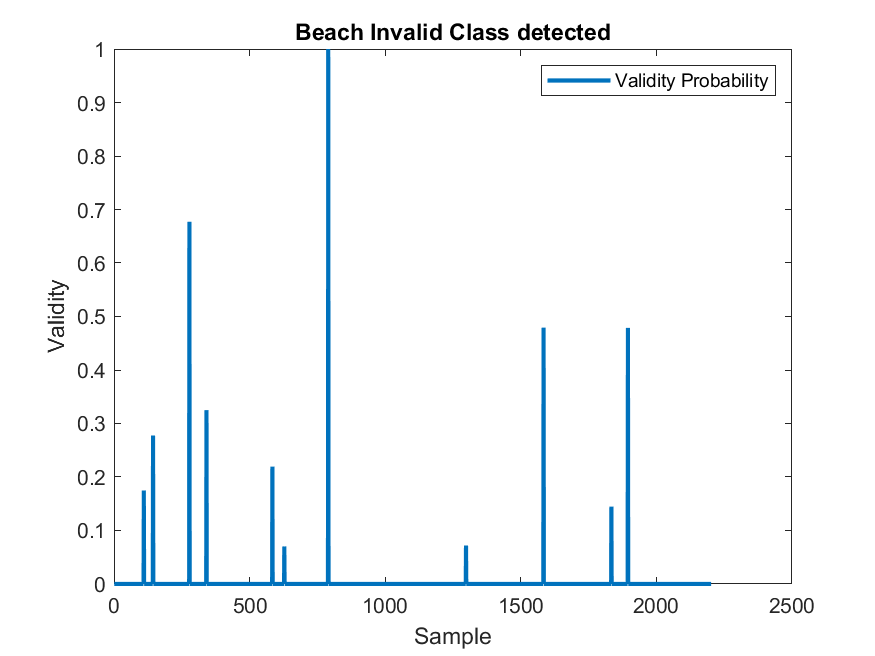}
	\includegraphics[width=0.3\linewidth]{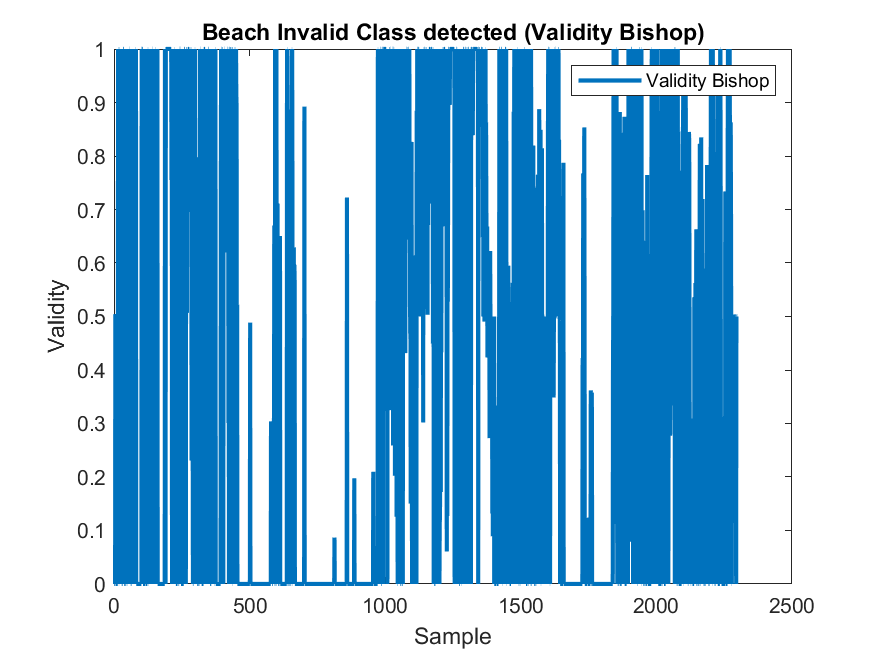}
	\caption{The first (top) visualization is a confusion matrix for the Beach Water Quality classifier accuracy. The bottom right corner of the confusion matrix shows the overall accuracy of the classifier. The three central plots  show the validity signal per sample for correctly classified examples (left kNN, center Probability, and right re-implementation of \cite{bishop1993neural}). For the incorrect classified examples the validity signal per sample is shown in the three plots at the bottom in the same order.}
	\label{fig:class2}
\end{figure}

\begin{figure}[htb]
	\centering
	\includegraphics[width=0.8\linewidth]{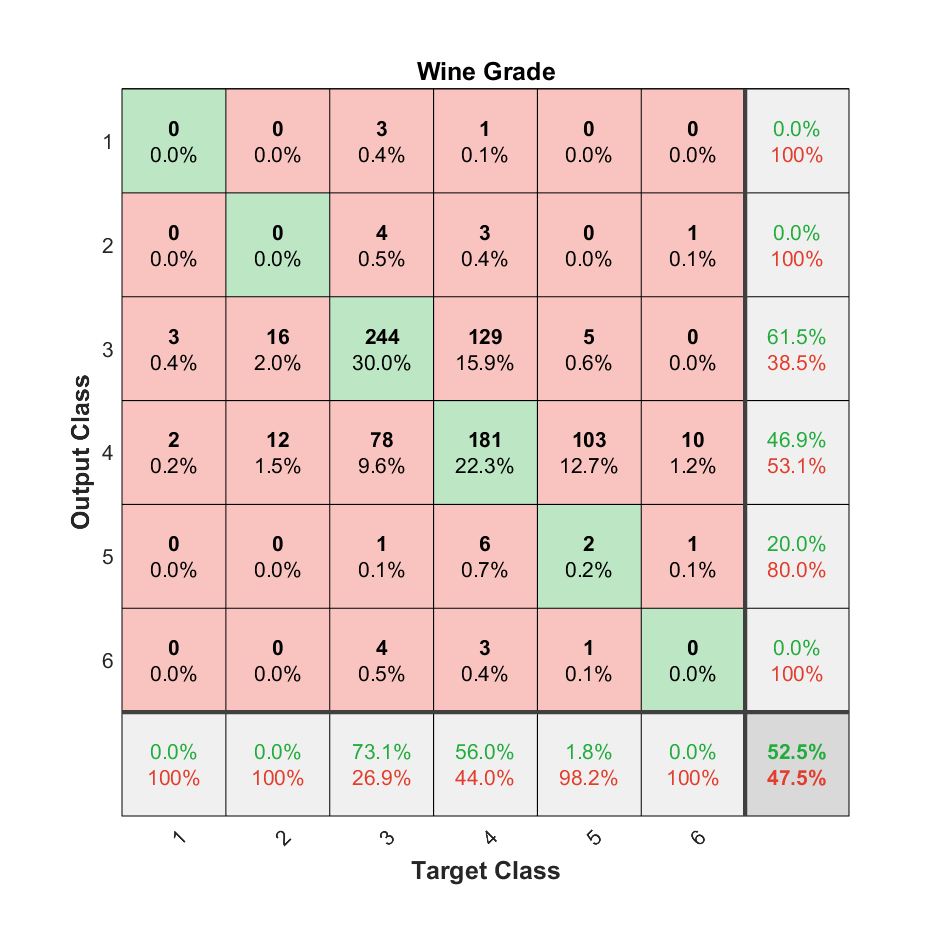}
	
	\includegraphics[width=0.3\linewidth]{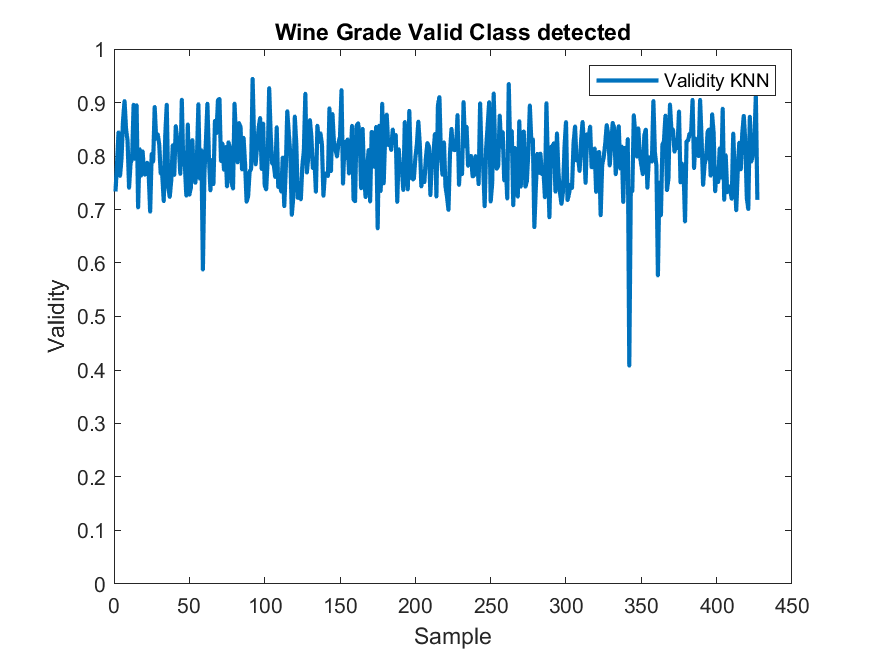}
	\includegraphics[width=0.3\linewidth]{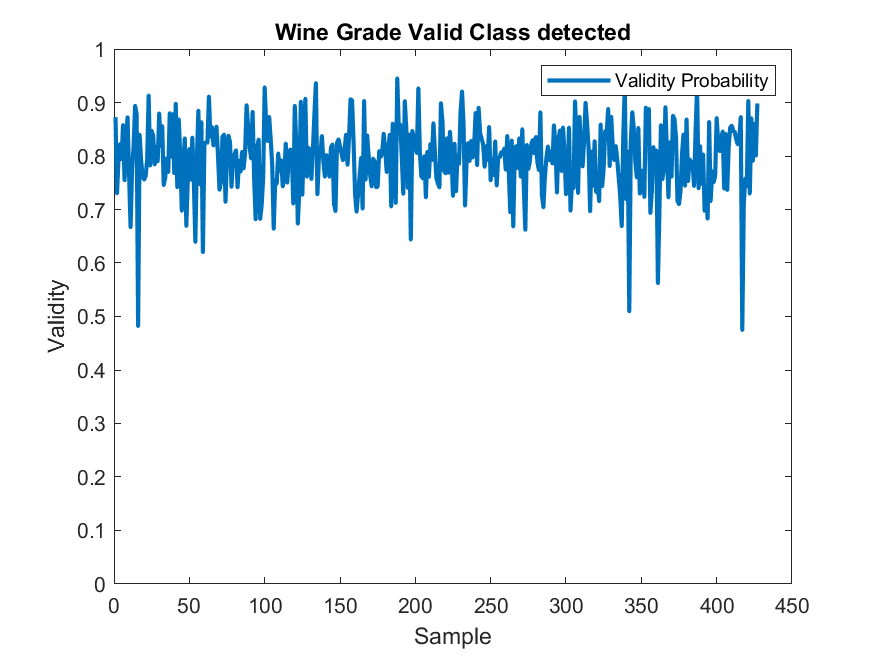}
	\includegraphics[width=0.3\linewidth]{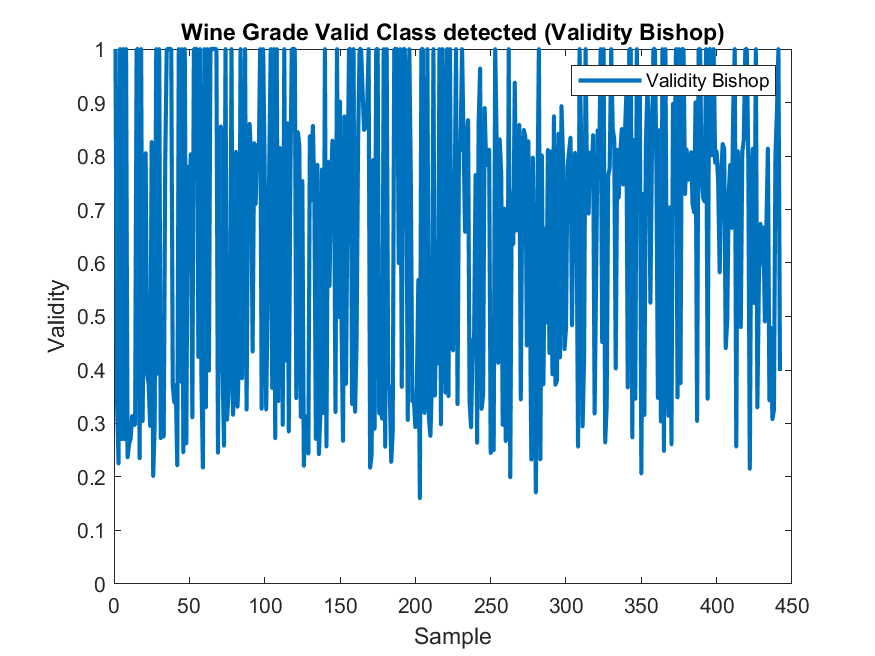}
	
	\includegraphics[width=0.3\linewidth]{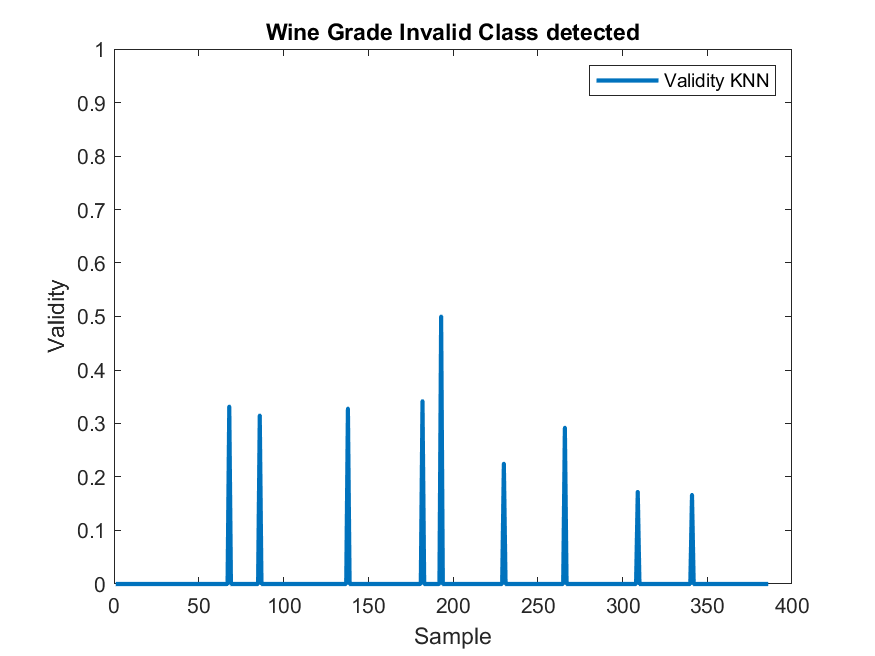}
	\includegraphics[width=0.3\linewidth]{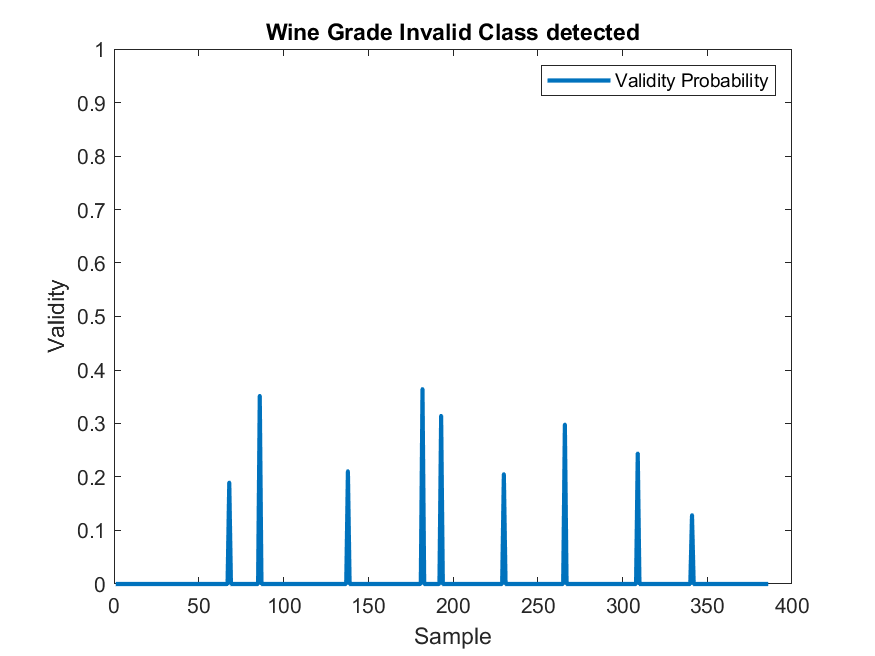}
	\includegraphics[width=0.3\linewidth]{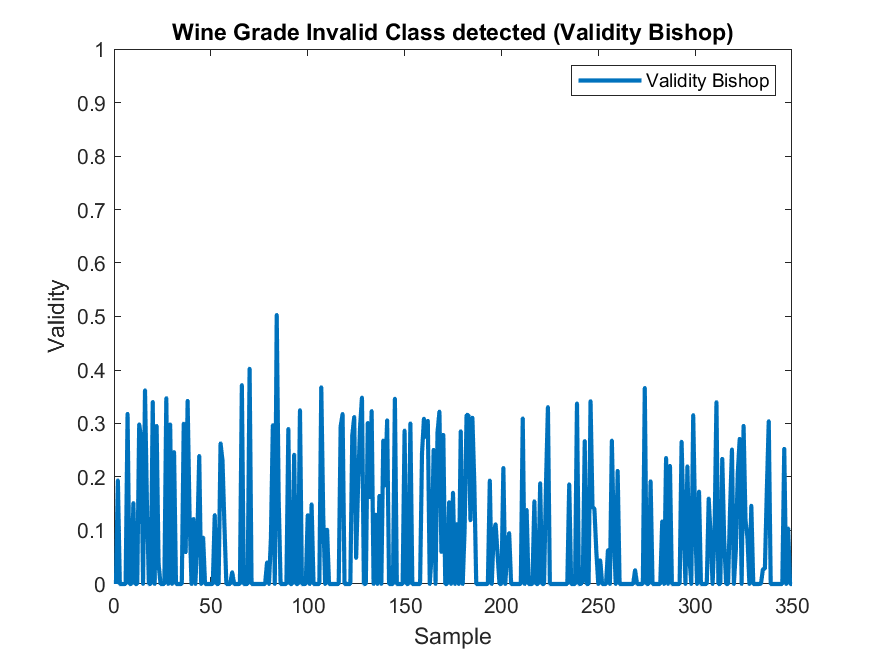}
	\caption{The first (top) visualization is a confusion matrix for the Red Wine Quality classifier accuracy. The bottom right corner of the confusion matrix shows the overall accuracy of the classifier. The three central plots  show the validity signal per sample for correctly classified examples (left kNN, center Probability, and right re-implementation of \cite{bishop1993neural}). For the incorrect classified examples the validity signal per sample is shown in the three plots at the bottom in the same order.}
	\label{fig:class3}
\end{figure}

Figure~\ref{fig:class1},~\ref{fig:class2} and \ref{fig:class3} show the classification results as well as the validation signal. The classification results are displayed as a confusion matrix for each data set (top plot). The three central plots show the validation signal on the correctly recognized classes per algorithm (left kNN, center Probability, and right re-implementation of \cite{bishop1993neural}). Here it is desirable that the validation signal is as high as possible. As can be seen in all plots, the signals are high for both the kNN and for our conditional probability approach, but not for the \cite{bishop1993neural} in Figure~\ref{fig:class1} and \ref{fig:class3}. Here the validation signal of the re-implementation has a significant impact on the classification accuracy.


Figure~\ref{fig:class1} shows that for some correctly recognized classes the validation signal of the kNN and Probability approach slips below a validity of 60\%. This reduces the accuracy of the classifier.  In contrast to the correctly recognized classes are the incorrectly recognized classes, which are shown in the three plots on the bottom. Here it is desirable that the validation signal is as low as possible. As can be seen in Figure~\ref{fig:class1}, this is the case for all wrongly detected classes and the validation signal of the two proposed methods, and worse for the re-implementation of \cite{bishop1993neural}.

In Figure~\ref{fig:class2} this looks a little different. Here we are trying to classify the beach based on the water sensor values. If we look at the validation signal in the bottom three plots, we see that for the kNN three entries and for the conditional probability approach two entries are wrongly considered valid. This is due to the similarity of the data, as it is possible that on some days the water quality on the beaches does not differ. The reason kNN scores worse is that the data are very unbalanced, as can be seen from the confusion matrix in Figure~\ref{fig:class2}. In the case of correctly classified data, only one entry is considered invalid by our conditional probability approach and three entries by the kNN. This has the same reasons as for the incorrectly classified data. \cite{bishop1993neural} provides worse results in both cases.

For the Wine Quality Classification in Figure~\ref{fig:class3}, the confusion matrix again shows very unbalanced data. The validation, however, works very well for the wrongly classified data, where all three approaches do not consider a wrongly classified entry as valid. In the case of correctly classified data, four entries are considered invalid by the conditional probability approach and three entries are marked as invalid by the kNN approach.

\begin{figure}[htb]
	\centering
	\includegraphics[width=0.48\linewidth]{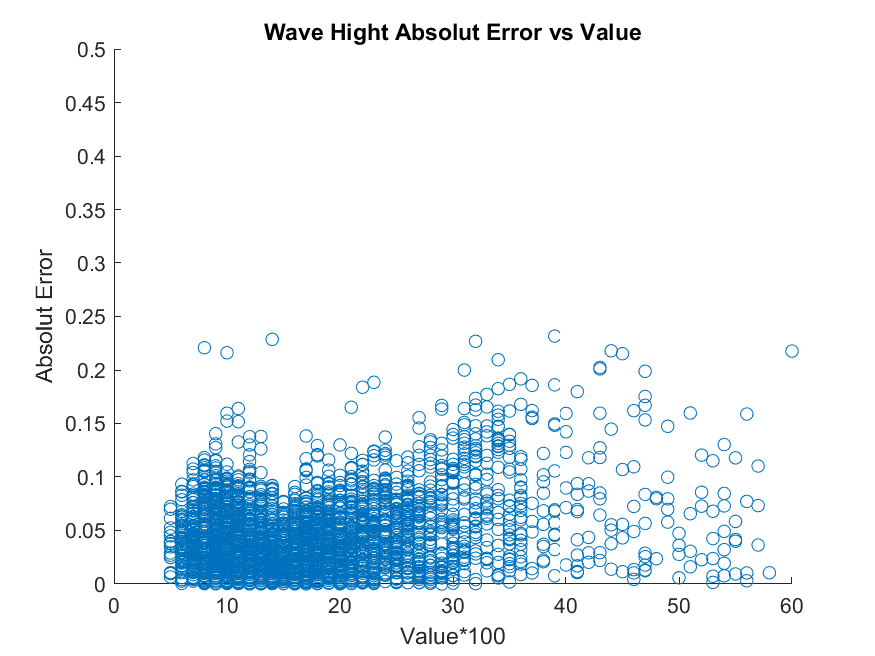}
	\includegraphics[width=0.48\linewidth]{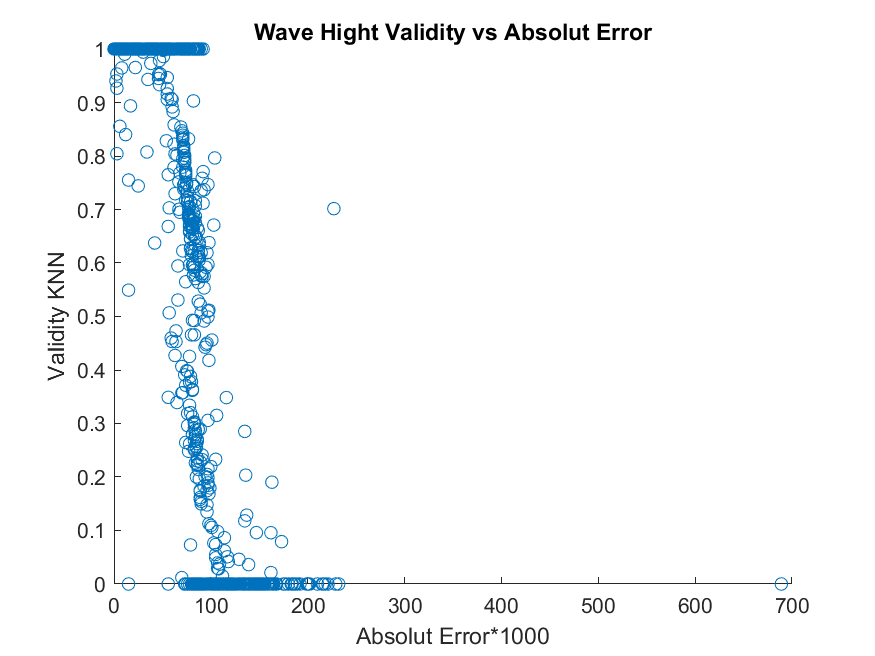}
	
	\includegraphics[width=0.48\linewidth]{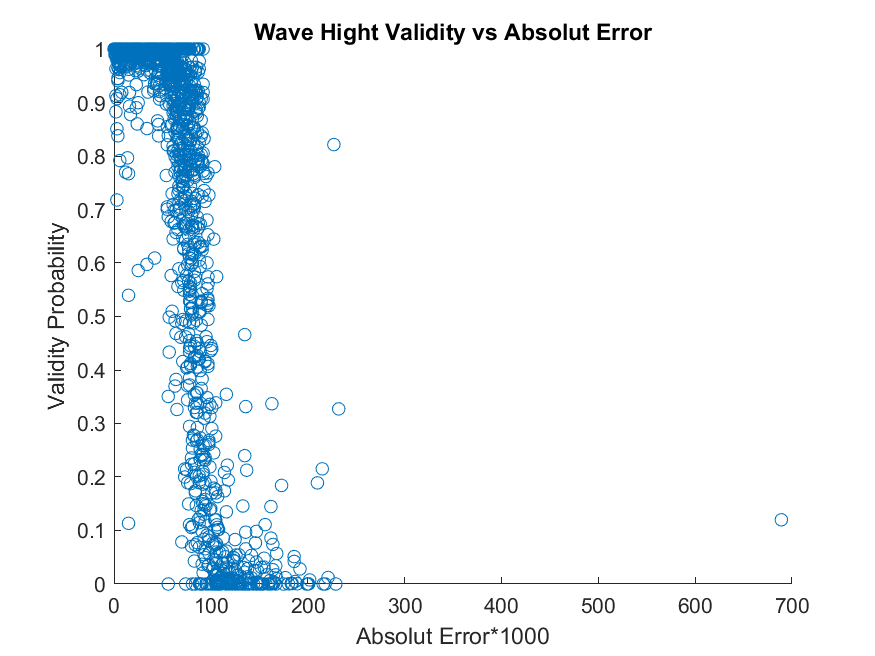}
	\includegraphics[width=0.48\linewidth]{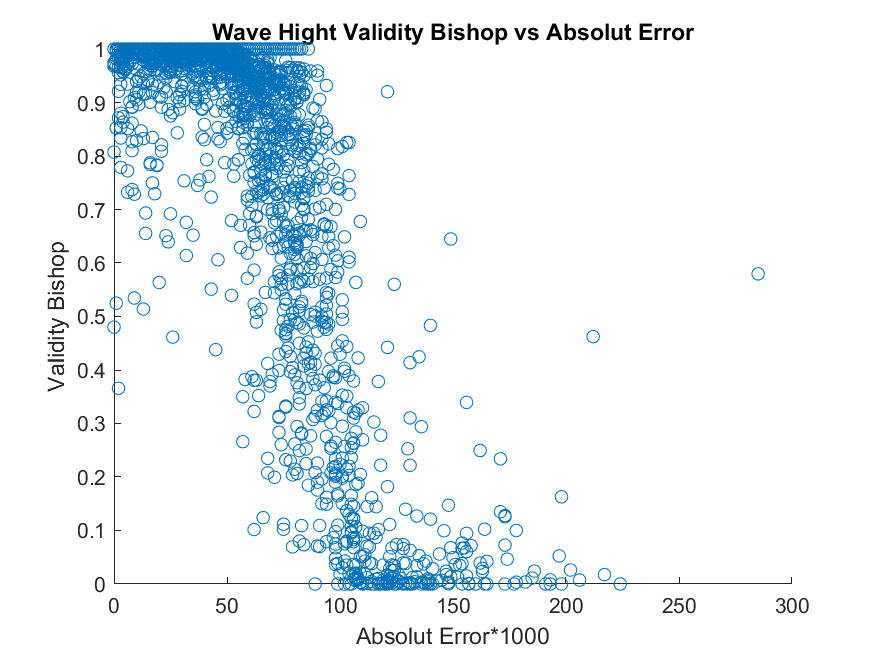}
	\caption{The first (top left) visualization shows the absolute error (y axis) related to the target waive height value (x axis). The three following plots (kNN, Probability, and re-implementation of \cite{bishop1993neural} in the same order) show the validity signal (y axis) to the error of the prediction (x axis). This means that the validity (y axis) should decrease, the higher the value on the x-axis will be.}
	\label{fig:reg1}
\end{figure}

\begin{figure}[htb]
	\centering
	\includegraphics[width=0.48\linewidth]{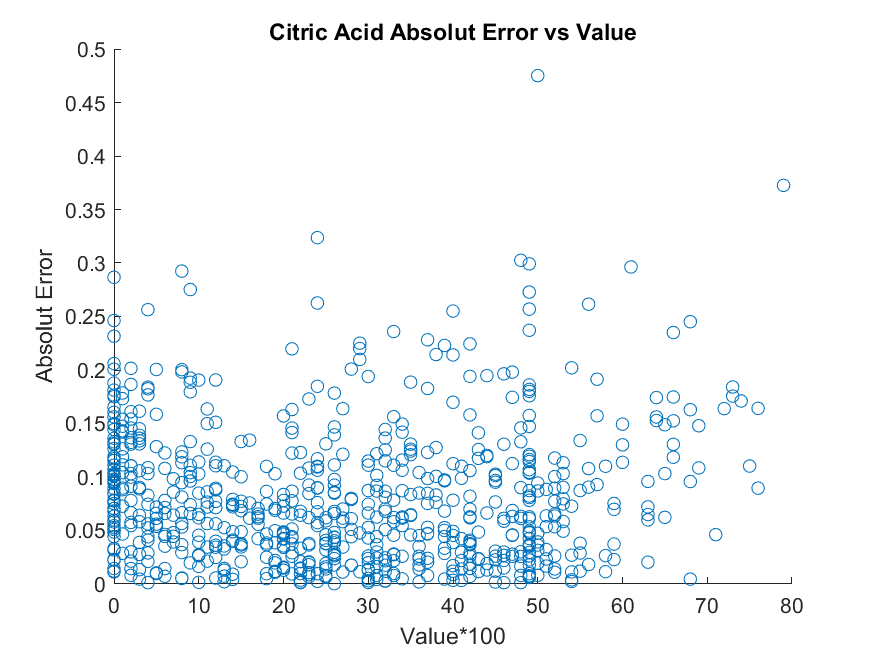}
	\includegraphics[width=0.48\linewidth]{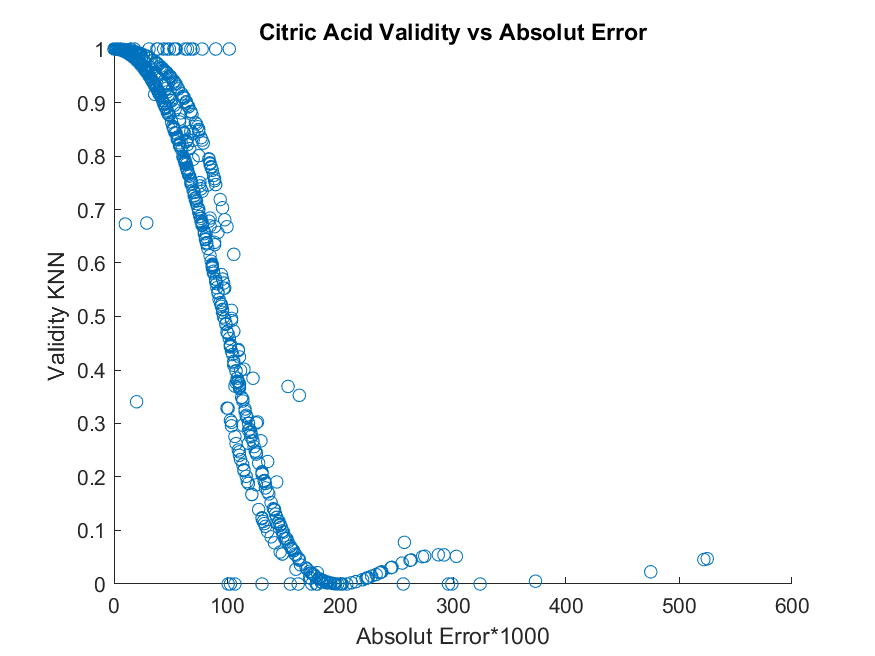}
	
	\includegraphics[width=0.48\linewidth]{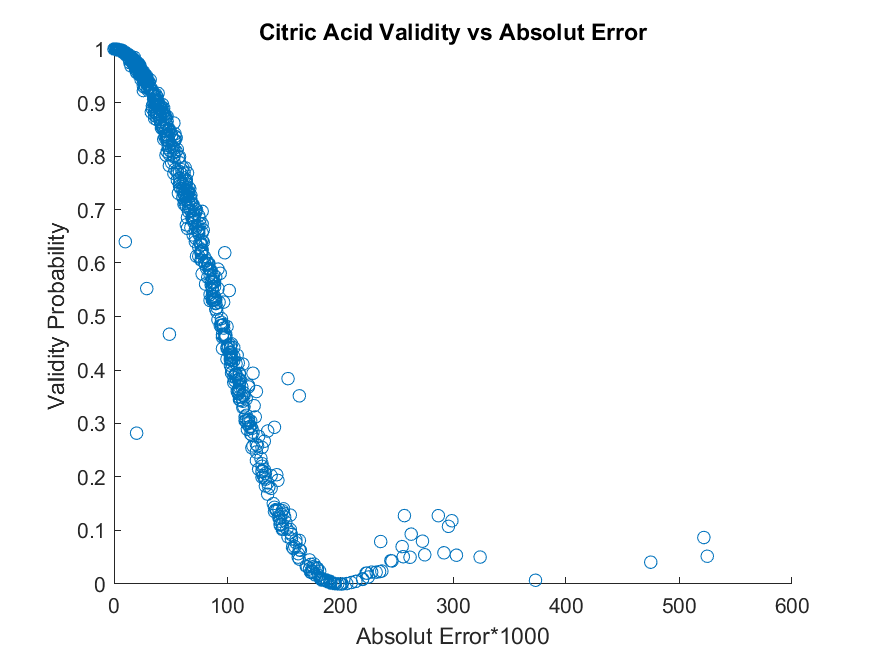}
	\includegraphics[width=0.48\linewidth]{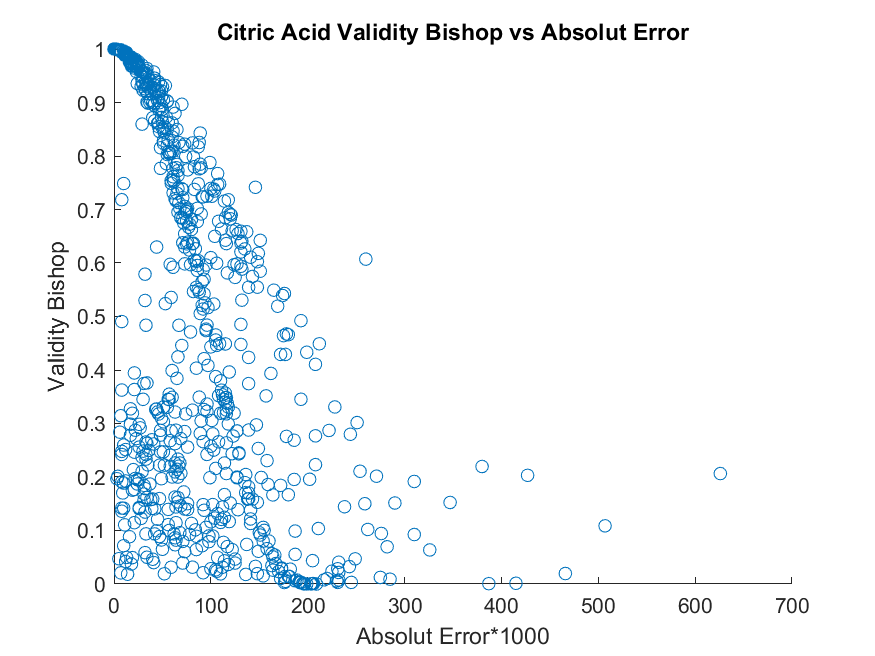}
	\caption{The first (top left) visualization shows the absolute error (y axis) related to the target ciric acid value (x axis). The three following plots (kNN, Probability, and re-implementation of \cite{bishop1993neural} in the same order) show the validity signal (y axis) to the error of the prediction (x axis). This means that the validity (y axis) should decrease, the higher the value on the x-axis will be.}
	\label{fig:reg2}
\end{figure}

For the evaluation of the regression, we have decided to quantify to 10 bins with an enlargement of the target value by a factor of 10. The maximum target value after enlargement is assumed to be 10. This applies to the regression of the wave height as well as to the citric acid. We used the value of 10 for quantification since the wave height are fixed values between 1 to 10 and the ciric acid value range was from 4,6 to 15,9. The enlargement factor of 10 was selected since the numerical accuracy for the ciric acid value is $10^{-1}$. The results are shown in Figures~\ref{fig:reg1} and \ref{fig:reg2}. The first plot shows the error distribution (y-axis) regarding the true magnitude on the x-axis (Figure~\ref{fig:reg1} the wave height and Figure~\ref{fig:reg2} the citric acid content). The three subsequent plots show the validation signal of the kNN, the conditional probabilities, and the re-implementation of \cite{bishop1993neural}. As can be seen, the validation signal decreases with increasing error, which shows the desired behaviour of our approach. To express this numerically we accept an error of 0.1 (corresponding to 100 in Figure~\ref{fig:reg1} and Figure~\ref{fig:reg2} of the second, third, and fourth plots) since we have quantized the values to 10 and take 10 as the maximum value. If we take this error as a threshold value, we get that for citric acid kNN recognized 6.7\% of all correct results as invalid and interpreted 1.1\% of all wrong values as valid. For the conditional probability approach, 14.2\% of the correct values are marked as invalid and 0\% of the incorrect values as valid. In comparison to this the re-implementation of \cite{bishop1993neural} marked 42.7\% of the correct values as invalid and 13.2\% of the incorrect values as valid. For the wave height with the same threshold value of 0.1, kNN marks 4.2\% of the correctly recognized heights as invalid and 1.2\% of the incorrectly recognized heights as valid. The conditional probability approach marks 4.3\% of the correctly detected heights as invalid and 1.2\% of the incorrectly detected heights as valid. The re-implementation of \cite{bishop1993neural} marked 4.5\% of the correct values as invalid and 4.2\% of the incorrect values as valid.

\section{Conclusion}
In this work we have shown how to transform a multi class and regression problem into a two class problem with the classes valid or invalid. This two class problem is easier to solve and can be used to validate the result of a machine learning algorithm. For this purpose, an explanatory algorithm is necessary why we have evaluated a KNN for this purpose in this work. Since a KNN with a constantly growing amount of training data requires more and more memory and the computing time is increasing because more comparisons have to be made, we presented an alternative solution using probabilities. Both methods were evaluated on three public datasets where two regression problems and three classification problems were considered for a variety of machine learning algorithms and compared against a re-implementation of a state-of-the-art algorithm~\cite{bishop1993neural}. Our results show that our approach provides a high quality validity signal and works on a variety of problems. Both presented approaches can be used online and the probabilistic approach can be executed on a microcontroller even with large data sets. Together with the explanatory power of validation based on the training data, our approach represents a step towards machine learning algorithms in critical areas of application. Future work will go in the direction of sensor data acquisition regarding critical areas to further evaluate and validate our method for a real application.

\bibliographystyle{acm}
\bibliography{IEEEbibfile}

\begin{thebibliography}{10}

\bibitem{alsheikh2014machine}
{\sc Alsheikh, M.~A., Lin, S., Niyato, D., and Tan, H.-P.}
\newblock Machine learning in wireless sensor networks: Algorithms, strategies,
  and applications.
\newblock {\em IEEE Communications Surveys \& Tutorials 16}, 4 (2014).

\bibitem{Bahmani2016}
{\sc Bahmani, H., Fuhl, W., Gutierrez, E., Kasneci, G., Kasneci, E., and Wahl,
  S.}
\newblock Feature-based attentional influences on the accommodation response.
\newblock In {\em Vision Sciences Society Annual Meeting Abstract\/} (2016).

\bibitem{beam2018big}
{\sc Beam, A.~L., and Kohane, I.~S.}
\newblock Big data and machine learning in health care.
\newblock {\em Jama 319}, 13 (2018), 1317--1318.

\bibitem{bell2016associations}
{\sc Bell, R.~P., Reed, S.~K., Schoonover, M.~J., Whitfield, C.~T., Yonezawa,
  Y., Maki, H., Pai, P.~F., and Keegan, K.~G.}
\newblock Associations of force plate and body-mounted inertial sensor
  measurements for identification of hind limb lameness in horses.
\newblock {\em American Journal of Veterinary Research 77}, 4 (2016), 337--345.

\bibitem{bibri2018iot}
{\sc Bibri, S.~E.}
\newblock The iot for smart sustainable cities of the future: An analytical
  framework for sensor-based big data applications for environmental
  sustainability.
\newblock {\em Sustainable cities and society 38\/} (2018), 230--253.

\bibitem{bishop1993neural}
{\sc Bishop, C.~M.}
\newblock Neural network validation: an illustration from the monitoring of
  multi-phase flows.
\newblock In {\em 1993 Third International Conference on Artificial Neural
  Networks\/} (1993), IET, pp.~41--45.

\bibitem{bishop1994novelty}
{\sc Bishop, C.~M.}
\newblock Novelty detection and neural network validation.
\newblock {\em IEEE Proceedings-Vision, Image and Signal processing 141}, 4
  (1994), 217--222.

\bibitem{breiman2001random}
{\sc Breiman, L.}
\newblock Random forests.
\newblock {\em Machine learning 45}, 1 (2001), 5--32.

\bibitem{bunel2020branch}
{\sc Bunel, R., Lu, J., Turkaslan, I., Kohli, P., Torr, P., and Mudigonda, P.}
\newblock Branch and bound for piecewise linear neural network verification.
\newblock {\em Journal of Machine Learning Research 21}, 2020 (2020).

\bibitem{bunel2018unified}
{\sc Bunel, R.~R., Turkaslan, I., Torr, P., Kohli, P., and Mudigonda, P.~K.}
\newblock A unified view of piecewise linear neural network verification.
\newblock In {\em Advances in Neural Information Processing Systems\/} (2018),
  pp.~4790--4799.

\bibitem{char2018implementing}
{\sc Char, D.~S., Shah, N.~H., and Magnus, D.}
\newblock Implementing machine learning in health care—addressing ethical
  challenges.
\newblock {\em The New England journal of medicine 378}, 11 (2018), 981.

\bibitem{cheng2018neural}
{\sc Cheng, C.-H., Diehl, F., Hinz, G., Hamza, Y., N{\"u}hrenberg, G., Rickert,
  M., Ruess, H., and Truong-Le, M.}
\newblock Neural networks for safety-critical applications—challenges,
  experiments and perspectives.
\newblock In {\em 2018 Design, Automation \& Test in Europe Conference \&
  Exhibition (DATE)\/} (2018), IEEE, pp.~1005--1006.

\bibitem{chowdhury2017protecting}
{\sc Chowdhury, M., Rahman, A., and Islam, R.}
\newblock Protecting data from malware threats using machine learning
  technique.
\newblock In {\em 2017 12th IEEE Conference on Industrial Electronics and
  Applications (ICIEA)\/} (2017), IEEE, pp.~1691--1694.

\bibitem{cortez2009modeling}
{\sc Cortez, P., Cerdeira, A., Almeida, F., Matos, T., and Reis, J.}
\newblock Modeling wine preferences by data mining from physicochemical
  properties.
\newblock {\em Decision Support Systems 47}, 4 (2009), 547--553.

\bibitem{cross2017use}
{\sc Cross, E.~S., Williams, L.~R., Lewis, D.~K., Magoon, G.~R., Onasch, T.~B.,
  Kaminsky, M.~L., Worsnop, D.~R., and Jayne, J.~T.}
\newblock Use of electrochemical sensors for measurement of air pollution:
  correcting interference response and validating measurements.
\newblock {\em Atmospheric Measurement Techniques 10}, 9 (2017), 3575.

\bibitem{czichos2018measurement}
{\sc Czichos, H.}
\newblock {\em Measurement, Testing and Sensor Technology}.
\newblock Springer, 2018.

\bibitem{dillmann2000learning}
{\sc Dillmann, R., Rogalla, O., Ehrenmann, M., Z{\"o}liner, R., and Bordegoni,
  M.}
\newblock Learning robot behaviour and skills based on human demonstration and
  advice: the machine learning paradigm.
\newblock In {\em Robotics Research}. Springer, 2000, pp.~229--238.

\bibitem{0320170}
{\sc Eivazi, S., Fuhl, W., and Kasneci, E.}
\newblock Towards intelligent surgical microscopes: Surgeons gaze and
  instrument tracking.
\newblock In {\em Proceedings of the 22st International Conference on
  Intelligent User Interfaces, IUI\/} (2017).

\bibitem{ACTNEURO2017}
{\sc Eivazi, S., Hafez, A., Fuhl, W., Afkari, H., Kasneci, E., Lehecka, M., and
  Bednarik, R.}
\newblock Optimal eye movement strategies: a comparison of neurosurgeons gaze
  patterns when using a surgical microscope.
\newblock {\em Acta Neurochirurgica\/} (2017).

\bibitem{032017}
{\sc Eivazi, S., Slupina, M., Fuhl, W., Afkari, H., Hafez, A., and Kasneci, E.}
\newblock Towards automatic skill evaluation in microsurgery.
\newblock In {\em Proceedings of the 22st International Conference on
  Intelligent User Interfaces, IUI 2017\/} (03 2017), ACM.

\bibitem{farrar2012structural}
{\sc Farrar, C.~R., and Worden, K.}
\newblock {\em Structural health monitoring: a machine learning perspective}.
\newblock John Wiley \& Sons, 2012.

\bibitem{UMUAI2020FUHL}
{\sc Fuhl, W.}
\newblock From perception to action using observed actions to learn gestures.
\newblock {\em User Modeling and User-Adapted Interaction\/} (08 2020), 1--18.

\bibitem{C2019}
{\sc Fuhl, W., Bozkir, E., Hosp, B., Castner, N., Geisler, D., Santini, T.~C.,
  and Kasneci, E.}
\newblock Encodji: encoding gaze data into emoji space for an amusing scanpath
  classification approach.
\newblock In {\em Proceedings of the 11th ACM Symposium on Eye Tracking
  Research \& Applications\/} (2019), pp.~1--4.

\bibitem{ICMIW2019FuhlW1}
{\sc Fuhl, W., Castner, N., and Kasneci, E.}
\newblock Histogram of oriented velocities for eye movement detection.
\newblock In {\em International Conference on Multimodal Interaction Workshops,
  ICMIW\/} (2018).

\bibitem{ICMIW2019FuhlW2}
{\sc Fuhl, W., Castner, N., and Kasneci, E.}
\newblock Rule based learning for eye movement type detection.
\newblock In {\em International Conference on Multimodal Interaction Workshops,
  ICMIW\/} (2018).

\bibitem{FFAO2019}
{\sc Fuhl, W., Castner, N., Kübler, T.~C., Lotz, A., Rosenstiel, W., and
  Kasneci, E.}
\newblock Ferns for area of interest free scanpath classification.
\newblock In {\em Proceedings of the 2019 ACM Symposium on Eye Tracking
  Research \& Applications (ETRA)\/} (06 2019).

\bibitem{ICCVW2018FuhlW}
{\sc Fuhl, W., Castner, N., Zhuang, L., Holzer, M., Rosenstiel, W., and
  Kasneci, E.}
\newblock Mam: Transfer learning for fully automatic video annotation and
  specialized detector creation.
\newblock In {\em International Conference on Computer Vision Workshops,
  ICCVW\/} (2018).

\bibitem{ETRA2018FuhlW}
{\sc Fuhl, W., Eivazi, S., Hosp, B., Eivazi, A., Rosenstiel, W., and Kasneci,
  E.}
\newblock Bore: Boosted-oriented edge optimization for robust, real time remote
  pupil center detection.
\newblock In {\em Eye Tracking Research and Applications, ETRA\/} (2018).

\bibitem{NNETRA2020}
{\sc Fuhl, W., Gao, H., and Kasneci, E.}
\newblock Neural networks for optical vector and eye ball parameter estimation.
\newblock In {\em ACM Symposium on Eye Tracking Research \& Applications, ETRA
  2020\/} (01 2020), ACM.

\bibitem{VECETRA2020}
{\sc Fuhl, W., Gao, H., and Kasneci, E.}
\newblock Tiny convolution, decision tree, and binary neuronal networks for
  robust and real time pupil outline estimation.
\newblock In {\em ACM Symposium on Eye Tracking Research \& Applications, ETRA
  2020\/} (01 2020), ACM.

\bibitem{ICCVW2019FuhlW}
{\sc Fuhl, W., Geisler, D., Rosenstiel, W., and Kasneci, E.}
\newblock The applicability of cycle gans for pupil and eyelid segmentation,
  data generation and image refinement.
\newblock In {\em International Conference on Computer Vision Workshops,
  ICCVW\/} (11 2019).

\bibitem{WDTTWE062018}
{\sc Fuhl, W., Geisler, D., Santini, T., Appel, T., Rosenstiel, W., and
  Kasneci, E.}
\newblock Cbf:circular binary features for robust and real-time pupil center
  detection.
\newblock In {\em ACM Symposium on Eye Tracking Research \& Applications\/} (06
  2018).

\bibitem{EPIC2018FuhlW}
{\sc Fuhl, W., and Kasneci, E.}
\newblock Eye movement velocity and gaze data generator for evaluation,
  robustness testing and assess of eye tracking software and visualization
  tools.
\newblock In {\em Poster at Egocentric Perception, Interaction and Computing,
  EPIC\/} (2018).

\bibitem{ICMV2019FuhlW}
{\sc Fuhl, W., and Kasneci, E.}
\newblock Learning to validate the quality of detected landmarks.
\newblock In {\em International Conference on Machine Vision, ICMV\/} (11
  2019).

\bibitem{AAAIFuhlW}
{\sc Fuhl, W., Kasneci, G., Rosenstiel, W., and Kasneci, E.}
\newblock Training decision trees as replacement for convolution layers.
\newblock In {\em Conference on Artificial Intelligence, AAAI\/} (02 2020).

\bibitem{ROIGA2018}
{\sc Fuhl, W., Kübler, T.~C., Brinkmann, H., Rosenberg, R., Rosenstiel, W.,
  and Kasneci, E.}
\newblock Region of interest generation algorithms for eye tracking data.
\newblock In {\em Third Workshop on Eye Tracking and Visualization (ETVIS), in
  conjunction with ACM ETRA\/} (06 2018).

\bibitem{WTCDOWE052017}
{\sc Fuhl, W., Kübler, T.~C., Hospach, D., Bringmann, O., Rosenstiel, W., and
  Kasneci, E.}
\newblock Ways of improving the precision of eye tracking data: Controlling the
  influence of dirt and dust on pupil detection.
\newblock {\em Journal of Eye Movement Research 10}, 3 (05 2017).

\bibitem{ASAOIB2015}
{\sc Fuhl, W., Kübler, T.~C., Sippel, K., Rosenstiel, W., and Kasneci, E.}
\newblock Arbitrarily shaped areas of interest based on gaze density gradient.
\newblock In {\em European Conference on Eye Movements, ECEM 2015\/} (08 2015).

\bibitem{AGAS2018}
{\sc Fuhl, W., K{\"u}bler, T.~C., Santini, T., and Kasneci, E.}
\newblock Automatic generation of saliency-based areas of interest for the
  visualization and analysis of eye-tracking data.
\newblock In {\em VMV\/} (2018), pp.~47--54.

\bibitem{FCDGR2020FUHL}
{\sc Fuhl, W., Rong, Y., and Enkelejda, K.}
\newblock Fully convolutional neural networks for raw eye tracking data
  segmentation, generation, and reconstruction.
\newblock In {\em Proceedings of the International Conference on Pattern
  Recognition\/} (2020), pp.~0--0.

\bibitem{CAIP2019FuhlW}
{\sc Fuhl, W., Rosenstiel, W., and Kasneci, E.}
\newblock 500,000 images closer to eyelid and pupil segmentation.
\newblock In {\em Computer Analysis of Images and Patterns, CAIP\/} (11 2019).

\bibitem{WTDTE022017}
{\sc Fuhl, W., Santini, T., Geisler, D., Kübler, T.~C., and Kasneci, E.}
\newblock Eyelad: Remote eye tracking image labeling tool.
\newblock In {\em 12th Joint Conference on Computer Vision, Imaging and
  Computer Graphics Theory and Applications (VISIGRAPP 2017)\/} (02 2017).

\bibitem{WTDTWE092016}
{\sc Fuhl, W., Santini, T., Geisler, D., Kübler, T.~C., Rosenstiel, W., and
  Kasneci, E.}
\newblock Eyes wide open? eyelid location and eye aperture estimation for
  pervasive eye tracking in real-world scenarios.
\newblock In {\em ACM International Joint Conference on Pervasive and
  Ubiquitous Computing: Adjunct publication -- PETMEI 2016\/} (09 2016).

\bibitem{WTE032017}
{\sc Fuhl, W., Santini, T., and Kasneci, E.}
\newblock Fast and robust eyelid outline and aperture detection in real-world
  scenarios.
\newblock In {\em IEEE Winter Conference on Applications of Computer Vision
  (WACV 2017)\/} (03 2017).

\bibitem{CORR2017FuhlW1}
{\sc Fuhl, W., Santini, T., and Kasneci, E.}
\newblock Fast camera focus estimation for gaze-based focus control.
\newblock In {\em CoRR\/} (2017).

\bibitem{fuhl2018simarxiv}
{\sc Fuhl, W., Santini, T., Kuebler, T., Castner, N., Rosenstiel, W., and
  Kasneci, E.}
\newblock Eye movement simulation and detector creation to reduce laborious
  parameter adjustments.
\newblock {\em arXiv preprint arXiv:1804.00970\/} (2018).

\bibitem{WTCDAHKSE122016}
{\sc Fuhl, W., Santini, T., Reichert, C., Claus, D., Herkommer, A., Bahmani,
  H., Rifai, K., Wahl, S., and Kasneci, E.}
\newblock Non-intrusive practitioner pupil detection for unmodified microscope
  oculars.
\newblock {\em Elsevier Computers in Biology and Medicine 79\/} (12 2016),
  36--44.

\bibitem{DWTE022017}
{\sc Geisler, D., Fuhl, W., Santini, T., and Kasneci, E.}
\newblock Saliency sandbox: Bottom-up saliency framework.
\newblock In {\em 12th Joint Conference on Computer Vision, Imaging and
  Computer Graphics Theory and Applications (VISIGRAPP 2017)\/} (02 2017).

\bibitem{hagan1994training}
{\sc Hagan, M.~T., and Menhaj, M.~B.}
\newblock Training feedforward networks with the marquardt algorithm.
\newblock {\em IEEE transactions on Neural Networks 5}, 6 (1994), 989--993.

\bibitem{hart1968condensed}
{\sc Hart, P.}
\newblock The condensed nearest neighbor rule (corresp.).
\newblock {\em IEEE transactions on information theory 14}, 3 (1968), 515--516.

\bibitem{hastie2009elements}
{\sc Hastie, T., Tibshirani, R., and Friedman, J.}
\newblock {\em The elements of statistical learning: data mining, inference,
  and prediction}.
\newblock Springer Science \& Business Media, 2009.

\bibitem{holzinger2016interactive}
{\sc Holzinger, A.}
\newblock Interactive machine learning for health informatics: when do we need
  the human-in-the-loop?
\newblock {\em Brain Informatics 3}, 2 (2016), 119--131.

\bibitem{hu2010qelar}
{\sc Hu, T., and Fei, Y.}
\newblock Qelar: A machine-learning-based adaptive routing protocol for
  energy-efficient and lifetime-extended underwater sensor networks.
\newblock {\em IEEE Transactions on Mobile Computing 9}, 6 (2010), 796--809.

\bibitem{hu2013mutantx}
{\sc Hu, X., Shin, K.~G., Bhatkar, S., and Griffin, K.}
\newblock Mutantx-s: Scalable malware clustering based on static features.
\newblock In {\em Presented as part of the 2013 $\{$USENIX$\}$ Annual Technical
  Conference ($\{$USENIX$\}$$\{$ATC$\}$ 13)\/} (2013), pp.~187--198.

\bibitem{huang2004credit}
{\sc Huang, Z., Chen, H., Hsu, C.-J., Chen, W.-H., and Wu, S.}
\newblock Credit rating analysis with support vector machines and neural
  networks: a market comparative study.
\newblock {\em Decision support systems 37}, 4 (2004), 543--558.

\bibitem{klingspor1997human}
{\sc Klingspor, V., Demiris, J., and Kaiser, M.}
\newblock Human-robot communication and machine learning.
\newblock {\em Applied Artificial Intelligence 11}, 7 (1997), 719--746.

\bibitem{krogh1995neural}
{\sc Krogh, A., and Vedelsby, J.}
\newblock Neural network ensembles, cross validation, and active learning.
\newblock In {\em Advances in neural information processing systems\/} (1995),
  pp.~231--238.

\bibitem{kuleshov2017mobile}
{\sc Kuleshov, A., Bernstein, A., and Burnaev, E.}
\newblock Mobile robot localization via machine learning.
\newblock In {\em International Conference on Machine Learning and Data Mining
  in Pattern Recognition\/} (2017), Springer, pp.~276--290.

\bibitem{kurd2003establishing}
{\sc Kurd, Z., and Kelly, T.}
\newblock Establishing safety criteria for artificial neural networks.
\newblock In {\em International Conference on Knowledge-Based and Intelligent
  Information and Engineering Systems\/} (2003), Springer, pp.~163--169.

\bibitem{kurd2007developing}
{\sc Kurd, Z., Kelly, T., and Austin, J.}
\newblock Developing artificial neural networks for safety critical systems.
\newblock {\em Neural Computing and Applications 16}, 1 (2007), 11--19.

\bibitem{lee2007application}
{\sc Lee, Y.-C.}
\newblock Application of support vector machines to corporate credit rating
  prediction.
\newblock {\em Expert Systems with Applications 33}, 1 (2007), 67--74.

\bibitem{liao2010anomaly}
{\sc Liao, Z., Yu, Y., and Chen, B.}
\newblock Anomaly detection in gps data based on visual analytics.
\newblock In {\em 2010 IEEE Symposium on Visual Analytics Science and
  Technology\/} (2010), IEEE, pp.~51--58.

\bibitem{luo2017deep}
{\sc Luo, C., Wu, D., and Wu, D.}
\newblock A deep learning approach for credit scoring using credit default
  swaps.
\newblock {\em Engineering Applications of Artificial Intelligence 65\/}
  (2017), 465--470.

\bibitem{meinshausen2006quantile}
{\sc Meinshausen, N.}
\newblock Quantile regression forests.
\newblock {\em Journal of Machine Learning Research 7}, Jun (2006), 983--999.

\bibitem{mohr2017personal}
{\sc Mohr, D.~C., Zhang, M., and Schueller, S.~M.}
\newblock Personal sensing: understanding mental health using ubiquitous
  sensors and machine learning.
\newblock {\em Annual review of clinical psychology 13\/} (2017), 23--47.

\bibitem{moller1990scaled}
{\sc M{\o}ller, M.~F.}
\newblock {\em A scaled conjugate gradient algorithm for fast supervised
  learning}.
\newblock Aarhus University, Computer Science Department, 1990.

\bibitem{muller2017organic}
{\sc M{\"u}ller-Schloer, C., and Tomforde, S.}
\newblock {\em Organic Computing-Technical Systems for Survival in the Real
  World}.
\newblock Springer, 2017.

\bibitem{naghibi2016gis}
{\sc Naghibi, S.~A., Pourghasemi, H.~R., and Dixon, B.}
\newblock Gis-based groundwater potential mapping using boosted regression
  tree, classification and regression tree, and random forest machine learning
  models in iran.
\newblock {\em Environmental monitoring and assessment 188}, 1 (2016), 44.

\bibitem{rodvold1999software}
{\sc Rodvold, D.~M.}
\newblock A software development process model for artificial neural networks
  in critical applications.
\newblock In {\em IJCNN'99. International Joint Conference on Neural Networks.
  Proceedings (Cat. No. 99CH36339)\/} (1999), vol.~5, IEEE, pp.~3317--3322.

\bibitem{schumann2003verification}
{\sc Schumann, J., Gupta, P., and Nelson, S.}
\newblock On verification \& validation of neural network based controllers.
\newblock {\em EANN’03\/} (2003).

\bibitem{serov2017cognitive}
{\sc Serov, A.}
\newblock Cognitive sensor technology for structural health monitoring.
\newblock {\em Procedia Structural Integrity 5\/} (2017), 1160--1167.

\bibitem{shamili2010malware}
{\sc Shamili, A.~S., Bauckhage, C., and Alpcan, T.}
\newblock Malware detection on mobile devices using distributed machine
  learning.
\newblock In {\em 2010 20th International Conference on Pattern Recognition\/}
  (2010), IEEE, pp.~4348--4351.

\bibitem{shcherbina2017accuracy}
{\sc Shcherbina, A., Mattsson, C.~M., Waggott, D., Salisbury, H., Christle,
  J.~W., Hastie, T., Wheeler, M.~T., and Ashley, E.~A.}
\newblock Accuracy in wrist-worn, sensor-based measurements of heart rate and
  energy expenditure in a diverse cohort.
\newblock {\em Journal of personalized medicine 7}, 2 (2017), 3.

\bibitem{sng2002strategy}
{\sc Sng, H., Gupta, G.~S., and Messom, C.~H.}
\newblock Strategy for collaboration in robot soccer.
\newblock In {\em Proceedings First IEEE International Workshop on Electronic
  Design, Test and Applications' 2002\/} (2002), IEEE.

\bibitem{tsai2010credit}
{\sc Tsai, C.-F., and Chen, M.-L.}
\newblock Credit rating by hybrid machine learning techniques.
\newblock {\em Applied soft computing 10}, 2 (2010), 374--380.

\bibitem{warmuth2006totally}
{\sc Warmuth, M.~K., Liao, J., and R{\"a}tsch, G.}
\newblock Totally corrective boosting algorithms that maximize the margin.
\newblock In {\em Proceedings of the 23rd international conference on Machine
  learning\/} (2006), pp.~1001--1008.

\bibitem{webster2018measurement}
{\sc Webster, J.~G., and Eren, H.}
\newblock {\em Measurement, Instrumentation, and Sensors Handbook: Two-Volume
  Set}.
\newblock CRC press, 2018.

\bibitem{wu2010study}
{\sc Wu, C.-H., Su, W.-H., and Ho, Y.-W.}
\newblock A study on gps gdop approximation using support-vector machines.
\newblock {\em IEEE Transactions on Instrumentation and Measurement 60}, 1
  (2010), 137--145.

\end{thebibliography}

\cleardoublepage

\end{document}